%% file: main.tex
\documentclass[10pt,twocolumn,letterpaper]{article}
\usepackage{cvpr} 
\usepackage{verbatim}
\usepackage{times}
\input{preamble}

\definecolor{cvprblue}{rgb}{0.21,0.49,0.74}
\usepackage[pagebackref,breaklinks,colorlinks,citecolor=cvprblue]{hyperref}

\title{SG-BEV: Satellite-Guided BEV Fusion for Cross-View Semantic Segmentation}
\author{
    \textrm{Junyan Ye\textsuperscript{\rm 1,2}}\thanks{This work was partially done during the internship at Shanghai Artificial Intelligence Laboratory.}\textrm{,}
    \textrm{Qiyan Luo\textsuperscript{\rm 1},} 
    \textrm{Jinhua Yu\textsuperscript{\rm 1},}
    \textrm{Huaping Zhong\textsuperscript{\rm 3},} 
    \\
    \textrm{Zhimeng Zheng\textsuperscript{\rm 2,4},}
    \textrm{Conghui He\textsuperscript{\rm 2,3},}
    \textrm{Weijia Li\textsuperscript{\rm 1}}\thanks{Corresponding author.}
    \\
    \textsuperscript{\rm 1}\textrm{Sun Yat-Sen University,} 
    \textsuperscript{\rm 2}\textrm{Shanghai AI Laboratory,}
    \textsuperscript{\rm 3}\textrm{SenseTime Research,}
    \textsuperscript{\rm 4}\textrm{Zhejiang University}
    \\
    {\tt\small
        \{jejy53, luoqy26, yujh56\}@mail2.sysu.edu.cn, 
        zhonghuaping@sensetime.com,
    }
    \\
    {\tt\small
        \{zhengzhimeng, heconghui\}@pjlab.org.cn,
        liweij29@mail.sysu.edu.cn
    } 
}

\usepackage{tabularx}
\newcolumntype{Y}{>{\centering\arraybackslash}X}

\begin{document}
\maketitle
 \input{sec/0_abstract}
 \input{sec/1_intro}
 \input{sec/2_related_work}
 \input{sec/3_methods}

 \input{sec/4_experiments}
 \input{sec/5_conclusion}
 \input{sec/Reference}
 \input{sec/X_suppl}

\end{document}

%% file: preamble.tex
\usepackage{multirow}
\usepackage[dvipsnames]{xcolor}


%% file: sec/0_abstract.tex

\begin{abstract}

This paper aims at achieving fine-grained building attribute segmentation in a cross-view scenario, i.e., using satellite and street-view image pairs. The main challenge lies in overcoming the significant perspective differences between street views and satellite views. In this work, we introduce SG-BEV, a novel approach for satellite-guided BEV fusion for cross-view semantic segmentation. To overcome the limitations of existing cross-view projection methods in capturing the complete building facade features, we innovatively incorporate Bird's Eye View (BEV) method to establish a spatially explicit mapping of street-view features. Moreover, we fully leverage the advantages of multiple perspectives by introducing a novel satellite-guided reprojection module, optimizing the uneven feature distribution issues associated with traditional BEV methods. Our method demonstrates significant improvements on four cross-view datasets collected from multiple cities, including New York, San Francisco, and Boston. On average across these datasets, our method achieves an increase in mIOU by 10.13\% and 5.21\% compared with the state-of-the-art satellite-based and cross-view methods. The code and datasets of this work will be released at \url{https://github.com/yejy53/SG-BEV}.

\end{abstract}

%% file: sec/1_intro.tex
\section{Introduction}
\label{sec:intro}

\begin{figure} [ht]
  \centering
  \includegraphics[width=8.2cm]{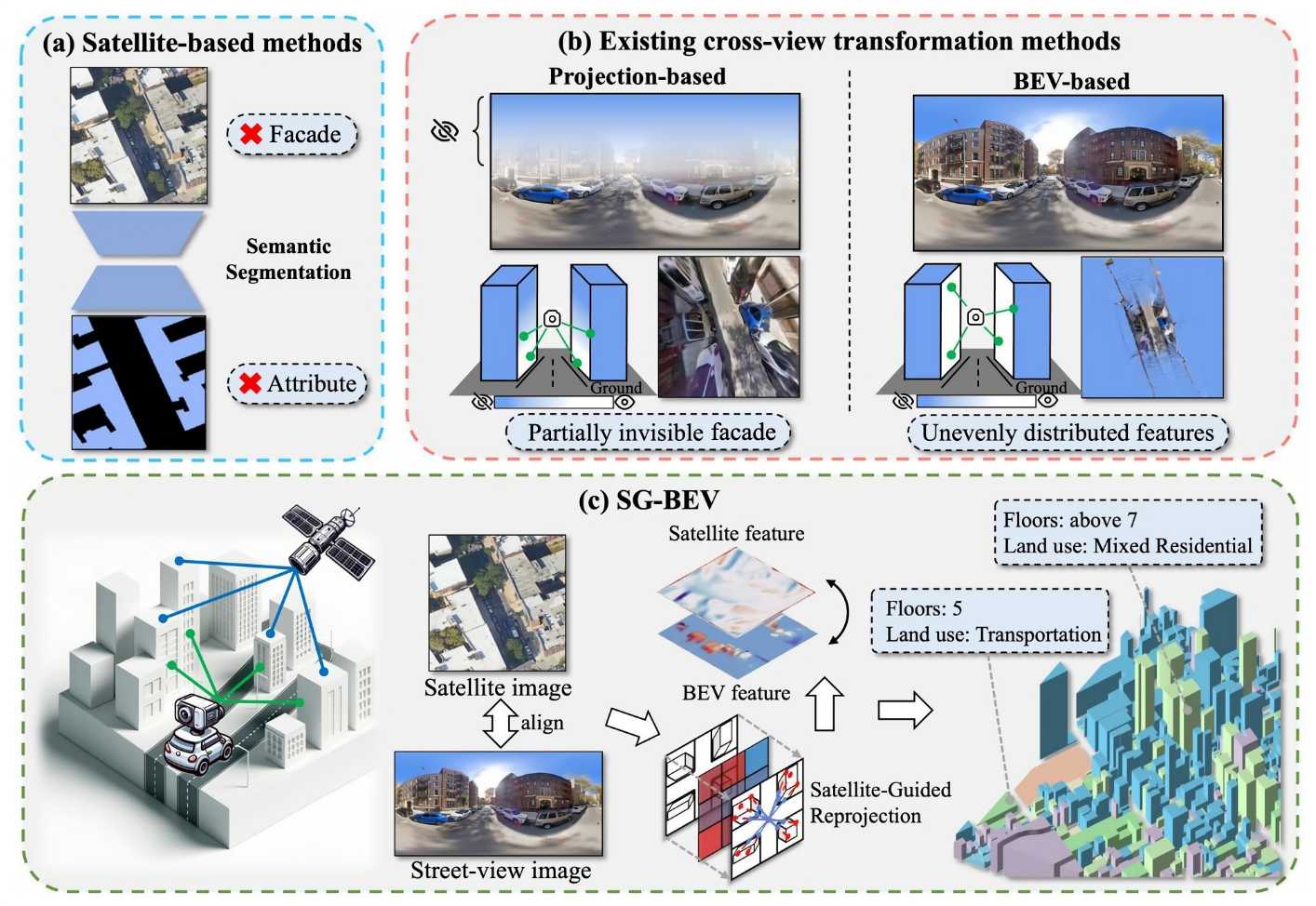}
  \caption{\textbf{Illustration of cross-view semantic segmentation of fine-grained building.} (a) Satellite imagery lacks information on building facades, making it difficult to distinguish detailed building attributes. (b) Existing cross-view transformation methods face issues with incomplete feature capture and uneven feature distribution. (c) Our method integrates satellite and street-view features to precisely segment building attributes and floor numbers.}
  \label{fig:figure}
\vspace{-10pt}
\end{figure}

\hspace*{1em}Fine-grained building attribute segmentation is a crucial task for urban planning, environment monitoring and residential management \cite{li2023joint,kemker2018algorithms, rottensteiner2012isprs}. Satellite images offers a comprehensive outline of building footprints, while street-view images contribute detailed facade features. Integrating these two types of data has demonstrated significant potential in achieving precise attribute segmentation of buildings \cite{workman2017unified,workman2022revisiting,srivastava2019understanding,lin2013cross}. In this paper, we focus on the cross-view semantic segmentation of fine-grained attributes using pairs of satellite and street-view images.

Previous studies on the semantic segmentation of fine-grained attributes for buildings or other terrestrial objects have predominantly relied on satellite images \cite{liu2023seeing,zheng2020foreground,li2023joint}.
However, as shown in Figure \ref{fig:figure}(a), the satellite perspective captures only the top and outline information, making it challenging to distinguish the fine-grained attribute differences between different buildings \cite{Li_2023_CVPR,workman2017unified,workman2022revisiting}. To address this issue, recent research has incorporated the street-view perspective to supplement facade information of buildings. The typical approach involves mapping street-view features to corresponding areas in the satellite view, thereby creating a link between satellite and street-view images \cite{srivastava2019understanding,workman2017unified,workman2022revisiting,feng2018urban}. However, existing approaches often reflect only general characteristics near the area, struggling to map street-view building features precisely to specific locations in the satellite view, leading to subpar performance in fine-grained attribute segmentation at the individual building level. To more effectively convey facade features from the street view for each building, exploring a novel cross-view feature mapping method that can continuously and precisely map street-view features to specific satellite view locations is necessary. The significant difference between street and satellite views poses a substantial challenge for precise cross-view feature mapping.

To effectively map and align features from street and satellite imagery, some current studies employed cross-view geometric projection methods \cite{shi2023boosting,wang2023fine,zhu2022transgeo}. 
However, these methods are more suitable for analyzing central ground areas, such as road regions for applications like image localization and driving planning \cite{yang2023parametric,shi2023boosting,wang2023fine}. 
In these cross-view mapping approaches, geometric projection is typically conducted through ground assumption and 360° panoramic mapping relations \cite{shi2023boosting,wang2023fine}. However, these methods often fail to effectively capture the facade features of taller buildings above the viewpoint, resulting in significant distortion of features away from the center area, as illustrated in Figure \ref{fig:figure}(b). 
This limitation leads to poor performance in comprehensively capturing the facade features of buildings, which is particularly evident in urban environments with high-rise structures. Such a constraint significantly restricts their capability in addressing cross-view fine-grained building attribute segmentation problems.


Bird's Eye View (BEV) methods represent another category of cross-view feature mapping, commonly used in autonomous driving or robot navigation \cite{li2022delving,li2022bevformer,peng2023bevsegformer,reiher2020sim2real,pan2020cross,philion2020lift,jiang2023polarformer,lang2019pointpillars,teng2024_360bev}. Compared with the geometric projection methods mentioned previously, BEV methods, leveraging 3D scene estimation, can capture more complete features of building facade. We plan to introduce the BEV approach to map street-view features onto satellite images, representing a novel attempt at fine-grained building segmentation tasks in cross-view scenarios.
However, as street-view images are captured from a ground perspective, they struggle to fully perceive the complete outline of building footprints.
When converting street-view images to BEV, features are mainly concentrated and stacked at the visible parts of roads and building wall edges \cite{li2022bevformer,philion2020lift} (as shown in Figure \ref{fig:figure}(b)). This results in uneven BEV feature distribution, limiting its performance in fusion with satellite features. We note that satellite images provide complete building contours, hence we introduce a Satellite-Guided Reprojection (SGR) module. This module relocates features from building edges to interiors, effectively addressing uneven feature distribution.

In this work, we introduce SG-BEV, a satellite-guided BEV fusion method for cross-view semantic segmentation. Unlike previous cross-view transformation approaches, our method establishes a clear spatial mapping relationship from the street-view to the satellite perspective, overcoming the limitations of geometric projection methods in capturing building facade features, and the uneven feature distribution issue of traditional BEV methods. 

Our main contributions are summarized as follows:


\begin{itemize}
  \item We innovatively apply BEV paradigm to the task of cross-view semantic segmentation of fine-grained building attributes, achieving a complete and continuous mapping of street-view features to a top-down perspective.
  \item We develop a Satellite-Guided Reprojection (SGR) module to further address the issue of features unevenly concentrated at the edges of buildings in BEV methods.
  \item Our method is evaluated on four cross-view datasets from cities including New York, San Francisco and Boston. On average across these datasets, it demonstrates an improvement of 10.13\% and 5.21\% in mIOU compared to the state-of-the-art satellite-based and cross-view methods.
\end{itemize}

%% file: sec/2_related_work.tex
\section{Related work}
\label{sec:related_work}

\begin{figure*}[t]
\centering
\includegraphics[width=\linewidth]{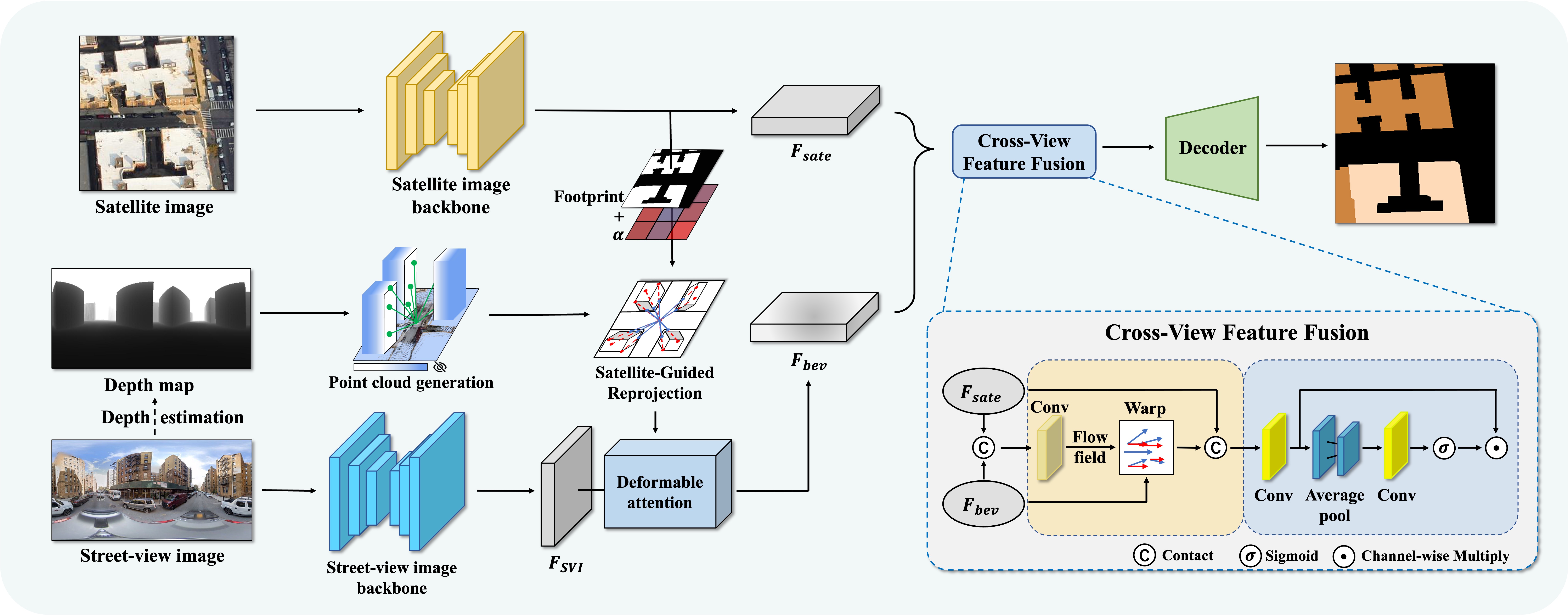}
\caption{
\textbf{Overview of our proposed SG-BEV framework.} In Satellite Feature Extraction branch, we extract features of input satellite imagery, meanwhile output building footprint segmentation results for further processing. In Street-View to BEV Conversion branch, we map street-view features to BEV space using estimated depth information combined with building footprints. In Cross-View Feature Fusion module, we align and fuse satellite features with BEV features to achieve fine-grained segmentation of building attributes.
}
\label{fig:pipeline}
\vspace{-10pt}
\end{figure*}


\subsection{Semantic Segmentation of Ground Objects}


\hspace*{1em}In the studies on semantic segmentation of ground objects, high-resolution satellite imagery has significantly contributed to the advancement of a variety of tasks, including urban road extraction \cite{chen2023semiroadexnet,bastani2018roadtracer}, land use classification \cite{xu2023rssformer}, and building extraction \cite{sikdar2023deepmao,chen2021cvcmff}. 
Prior research focused on buildings and other terrestrial objects, has primarily utilized satellite imagery as the main source of data, achieving notable results \cite{liu2023seeing,zheng2020foreground}. However, these approaches were somewhat limited in achieving fine-grained semantic segmentation, as they lacked facade information typically contained in street-view images. 
To address these limitations, Wojna et al. \cite{wojna2021holistic} introduced a method for projecting geometric attributes of buildings.
 Workman et al. utilized a backbone network to extract feature vectors representing the overall features of street views, using the spatial location of street-view images to diffuse these into the satellite view feature space \cite{workman2017unified}. Another approach involved creating a geospatial attention mechanism using distance and angle information, mapping street-view feature vectors onto the satellite view \cite{workman2022revisiting}. However, these feature mapping methods result in feature loss during the mapping process and sparse street-view features in the satellite perspective.
 Our method addresses these challenges by enabling accurate spatial mapping and efficient transfer of dense features from street views to the satellite view, thereby bridging the gap between different observational viewpoints.
\subsection{Cross-View Projection Methods}
\hspace*{1em}Cross-view projection methods play a crucial role in bridging the gap between different perspectives in image localization and driving planning \cite{shi2022beyond, toker2021coming, shi2020looking, shi2019spatial, lu2020geometry}. 
Techniques like polar transformations \cite{shi2019spatial} were employed by  to map features from satellite views to ground views. These transformations are crucial for tasks such as image retrieval and street-view generation. 
However, for the fine-grained building attribute segmentation from a top-down perspective that we aim to achieve, this method may not be suitable. 
In addition, several previous studies \cite{shi2023boosting, wang2023fine}, assumed a geometric relationship between the viewpoint and the ground plane to establish feature mapping relationship. 
However, these methods fail to capture features above the viewpoint and also create feature distortion away from the center area.
Our approach overcomes the incomplete feature mapping in cross-view geometric projection methods, achieving comprehensive mapping of building facade information.




\subsection{Bird's Eye View methods}



%

\hspace*{1em}The Bird's Eye View (BEV) methods have been widely used in autonomous driving and robot navigation \cite{li2022delving,peng2023bevsegformer,reiher2020sim2real,pan2020cross}, which are mainly for road area analysis and effective segmentation of targets like vehicles and lanes \cite{li2022hdmapnet,philion2020lift,jiang2023polarformer,peng2023bevsegformer}. 
The Lift, Splat, Shoot (LSS) \cite{philion2020lift} method mapped two-dimensional features to three-dimensional space by predicting depth distribution to acquire BEV features. 
BEVFormer \cite{li2022bevformer} startded from BEV queries and maps back to two-dimensional features for interaction. Additionally, BEVFormer enhanced the ability to capture features of tall objects by selecting multiple three-dimensional reference points along the \(Z\)-axis. 
However, street-view images fail to capture the complete outline of building footprints, leading to effective features being concentrated at the edges of depth-estimated dense areas (the building walls). In the LSS method, effective features were primarily concentrated at wall locations with the highest depth probabilities, while in BEVFormer, after average pooling along the \(Z\)-axis, features were also predominantly focused on the walls. 
The inconsistent distribution between BEV and satellite features, with strong features on building walls but sparse elsewhere, may degrade performance in subsequent tasks. 
Our designed Satellite-Guided Reprojection (SGR) module utilizes the footprint information provided by satellite imagery, combined with the estimated depth information, to guide the concentrated BEV features towards the interior of building footprints, effectively overcoming the issue of uneven BEV feature distribution.

%% file: sec/3_methods.tex
\section{Methods}
\label{sec:methods}



\hspace*{1em}As shown in Figure \ref{fig:pipeline}, this paper introduces a novel method for cross-view fine-grained attribute segmentation of buildings, named SG-BEV. In our comprehensive workflow, we employ two distinct branches to extract features from satellite and street-view images, respectively, and then merge them using a feature fusion module. In the satellite branch, we apply a backbone network to extract the satellite feature \(F_{\text{sat}}\) and output preliminary segmentation of the building footprint to guide the subsequent BEV features (Section \ref{section3.1}). In the street-view branch, we initially map using depth estimation information, then optimize the feature distribution with the Satellite-Guided Reprojection (SGR) module, , and finally produce BEV features \(F_{\text{bev}}\) with deformable attention \cite{li2022bevformer,wang2022detr3d} (Section \ref{section3.2}). In the Cross-View Feature Fusion module, we integrate \(F_{\text{sat}}\) with \(F_{\text{bev}}\) in a unified top-down view space (Section \ref{section3.3}). The decoder then processes these integrated features to produce fine-grained building attribute segmentation results, effectively capturing detailed attributes in the cross-view scenario.



\begin{figure}[t]
\centering
\includegraphics[width=\linewidth]{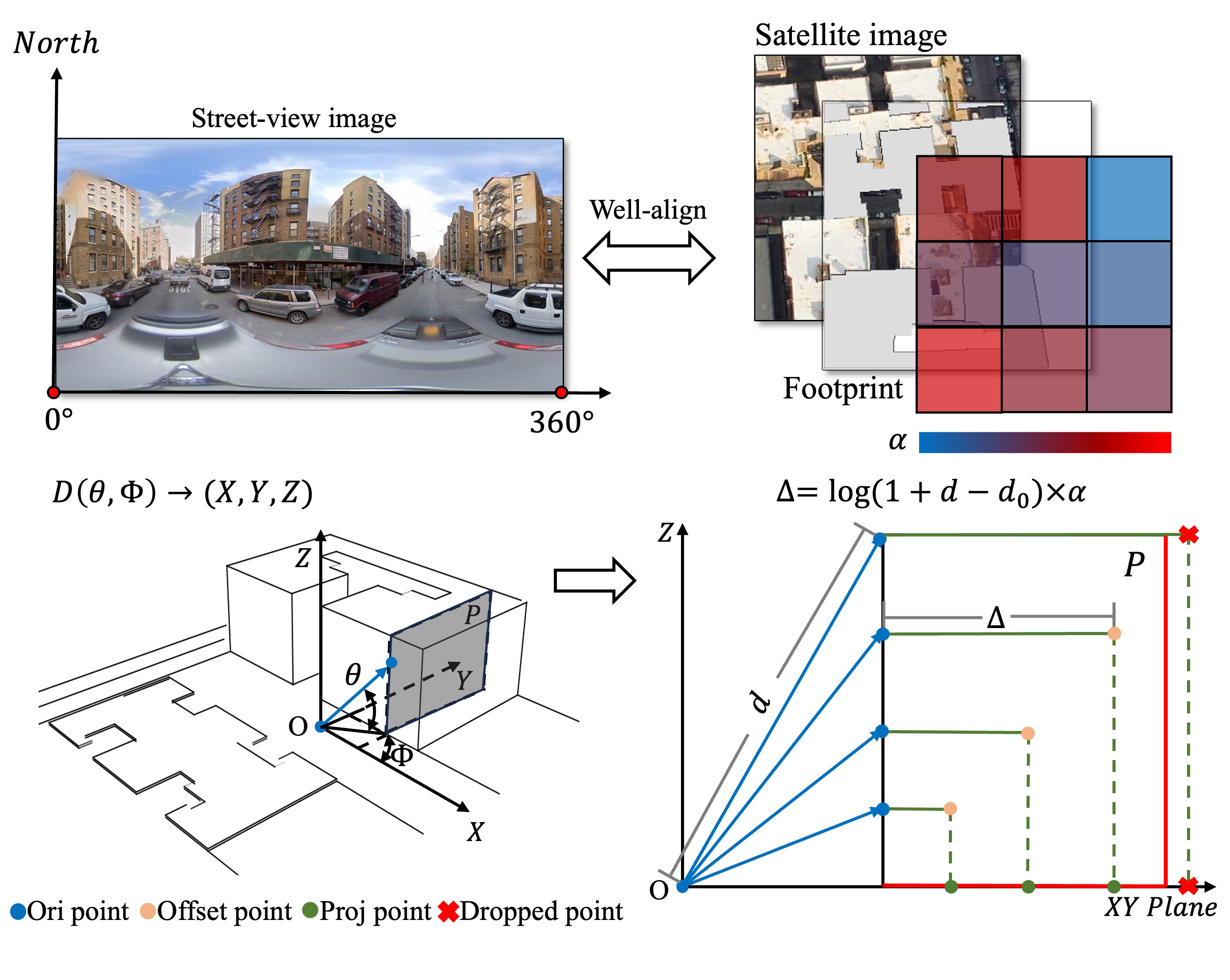}
\caption{\textbf{Illustration of Satellite-Guided Reprojection Module.} We utilize satellite features to generate building footprint information, followed by calculating \(\alpha\). Based on depth information \(d\) and \(\alpha\), we calculate magnitude of the offset \(\Delta\) to adjust the initial point cloud for uniform distribution within the building area and discard points that exceed the building's footprint.}
\vspace{-10pt}
\end{figure}

\subsection{Satellite Feature Extraction}
\label{section3.1}


\hspace*{1em}Satellite images provide a comprehensive outline of buildings from a top-down perspective, effectively compensating for the limitations of street-view images in perceiving the overall form of buildings and capturing areas obscured in street views. We deployed a feature extractor to process satellite images, thus obtaining the satellite features \(F_{\text{sat}}\). Since satellite images inherently offer a top-down perspective, they can be directly applied to subsequent Cross-View Feature Fusion. The final building contour features primarily originate from the satellite branch, which has a simpler structure. This shorter pathway design facilitates the backpropagation of loss.

Furthermore, the obtained satellite features will be processed through an additional decoder to produce segmentation results of building footprints, distinguishing between building and non-building areas. Our subsequent SGR module will leverage the advantages of multiple perspectives, guiding the sensible mapping of street-view features using the output building footprints to prevent BEV features from concentrating solely on building edges.

\subsection{Street-View to BEV Conversion}
\label{section3.2}
\hspace*{1em}\textbf{Initial Point Cloud Generation.} By utilizing established monocular depth estimation algorithms \cite{bhat2023zoedepth}, we are able to derive depth maps from street-view images. Based on depth estimation results and the geometric relationship of panoramic images, we obtained a three-dimensional XYZ estimation of the scene \cite{teng2024_360bev}. With this, we established an index mapping relationship between the 3D scene and panoramic images, facilitating the preliminary mapping of street-view features.

%


\begin{equation}
\Theta_{i,j} = \frac{i\pi}{H}, \quad \Phi_{i,j} = -\frac{2\pi j}{W} + \pi
\end{equation}
\begin{equation*}
i = \{0, \ldots, H-1\}, j = \{0, \ldots, W-1\}
\end{equation*}


Here, \( \Theta \) and \( \Phi \) are angle matrices of panoramic images with size \( H \times W \), consisting of two-dimensional Euler angular equivariant series, where i and j represent row and column numbers, respectively. Given the representation in spherical coordinate systems, each 3D point \( (X_{i,j}, Y_{i,j}, Z_{i,j}) \) in the camera coordinate system will be obtained through the calculation in Eq. (\ref{equ2}), where \( D_{i,j} \) is the panoramic depth information.
\[
X_{i,j} = D_{i,j} \cdot \sin(\Theta_{i,j}) \cdot \sin(\Phi_{i,j}),
\]
\begin{equation}
Y_{i,j} = D_{i,j} \cdot \cos(\Theta_{i,j}),
\label{equ2}
\end{equation}
\[
Z_{i,j} = D_{i,j} \cdot \sin(\Theta_{i,j}) \cdot \cos(\Phi_{i,j}).
\]

%

\textbf{Satellite-Guided Reprojection.} 
In our cross-view semantic segmentation task, we aim to reproject street-view features into building interiors completely and continuously with minimal alterations. We observe that while BEV features concentrate on building walls in the \(XY\) plane, they are dispersed and extended in the \(Z\)-axis, corresponding to street-view features from the facade base to the top. Using depth information \(d\) as a positively correlated offset factor in this context can factor effectively maintains the visual continuity and integrity of the facade features. With this method, features at the bottom of the building facade are guided closer to the center area, while the top features are relatively distanced from the center area.

Additionally, we extract building footprint information from satellite images to calculate the adjustment coefficient \(\alpha\) to control the intensity of the offset. Specifically, the satellite image is divided into a \(3 \times 3\) grid, and the proportion of building pixels in each grid is calculated to set the value of \(\alpha\). In our approach, a higher value of the adjustment coefficient \(\alpha\) indicates a larger footprint area of the building, necessitating a greater degree of offset. The specific magnitude of the offset \(\Delta\) is jointly constructed by  depth and \(\alpha\), as illustrated in Eq. (\ref{equ3}).

\begin{equation}
\Delta = \log(1 + d - d_0) \times \alpha 
\label{equ3}
\end{equation}

Here, \(d\) means depth, and \(d_0\) is a predefined hyper-parameter. \(\Delta\) is adjusted using a logarithmic function, aiming to reduce discontinuities in the point cloud on the same building facade caused by the rapid increase in depth with height. When  \(d < d_0\), no offset occurs. Next, we determine the direction of the point cloud offset. Considering that the camera is situated at the center, the point cloud should offset away from the center. The offset direction is determined by the following Eq. (\ref{equ4}):



\begin{equation}
\vec{D} =
\begin{bmatrix}
    X_{i,j} - c_x \\
    Y_{i,j} - c_y
\end{bmatrix}
\label{equ4}
\end{equation}

Here, \(c_x\) and \(c_y\) represent the coordinates of the center position. Utilizing the calculated offset direction and distance, we accordingly adjusted the \(XY\) coordinates of the point cloud. Our method combines information from satellite imagery and depth data to effectively optimize the distribution of the initial point cloud by shifting it as Eq. (\ref{equ5}).

\begin{equation}
\begin{bmatrix}
    X_{i,j}^{\prime}\\
    Y_{i,j}^{\prime}
\end{bmatrix} =
\Delta \cdot \vec{D} +
\begin{bmatrix}
    X_{i,j}\\
    Y_{i,j}
\end{bmatrix}
\label{equ5}
\end{equation}

Subsequently, we use the index information carried by the point cloud, which indicates the correspondence between the spatial positions in BEV space and the locations on the panorama. By integrating with deformable attention mechanisms \cite{li2022bevformer, wang2022detr3d}, we map the perspective features of the street-view onto the BEV plane. More visualization information can be found in the supplementary materials. More information on our street-view feature mapping visual comparison with other methods, and the \(\alpha\) parameter can be found in the supplementary materials.

\subsection{Cross-View Feature Fusion}
\label{section3.3}

 \hspace*{1em}By acquiring satellite features and street-view BEV features as described above, we have unified features from two different views under a top-down perspective. Recognizing the challenges posed by depth estimation errors and positional inaccuracies, our Cross-View Feature Fusion model first addresses aligning these diverse features \cite{dong2022SuperFusion}. The alignment process begins by generating a 2D flow field \(\Omega\), which is calculated based on the spatial discrepancies between \(F_\text{sat}\) and \(F_\text{bev}\). This involves using convolutional layers to predict coordinate offsets. The calculated flow field \(\Omega\) is then used to warp \(F_\text{bev}\) to align with \(F_\text{sat}\), which can be mathematically formulated as:

\begin{equation}
F_{\text{aligned}} = [\text{Warp}( F_{\text{bev}} , \Omega ), F_{\text{sat}}]
\end{equation}
where \( [ \cdot, \cdot ] \) denotes the concatenation along the channel dimension, \(\text{Warp}\) is a function applying the calculated offsets using bilinear interpolation, ensuring the spatial alignment of the features.  \(\Omega\) is the deformation field predicted by a convolutional network.

The next step is to the integration of the feature sets. We initiate a spatial fusion process by applying a  \( 3 \times 3 \) convolution layer, crucial for enhancing the spatial representation of the features. The output of this layer serves as the input for our adaptive integration function. As illustrated in Figure \ref{fig:pipeline}, the refined features from the convolution layer are first globally averaged pooled and then employ a linear transformation. This transformation is implemented via a \( 1 \times 1 \) convolution. Our cross-view feature fusion module captures essential information from both satellite and BEV features while minimizing their alignment errors.



%% file: sec/4_experiments.tex
\section{Experiments}
\label{sec:experiments}

\begin{figure*}[htbp]
\centering
\includegraphics[width=\linewidth]{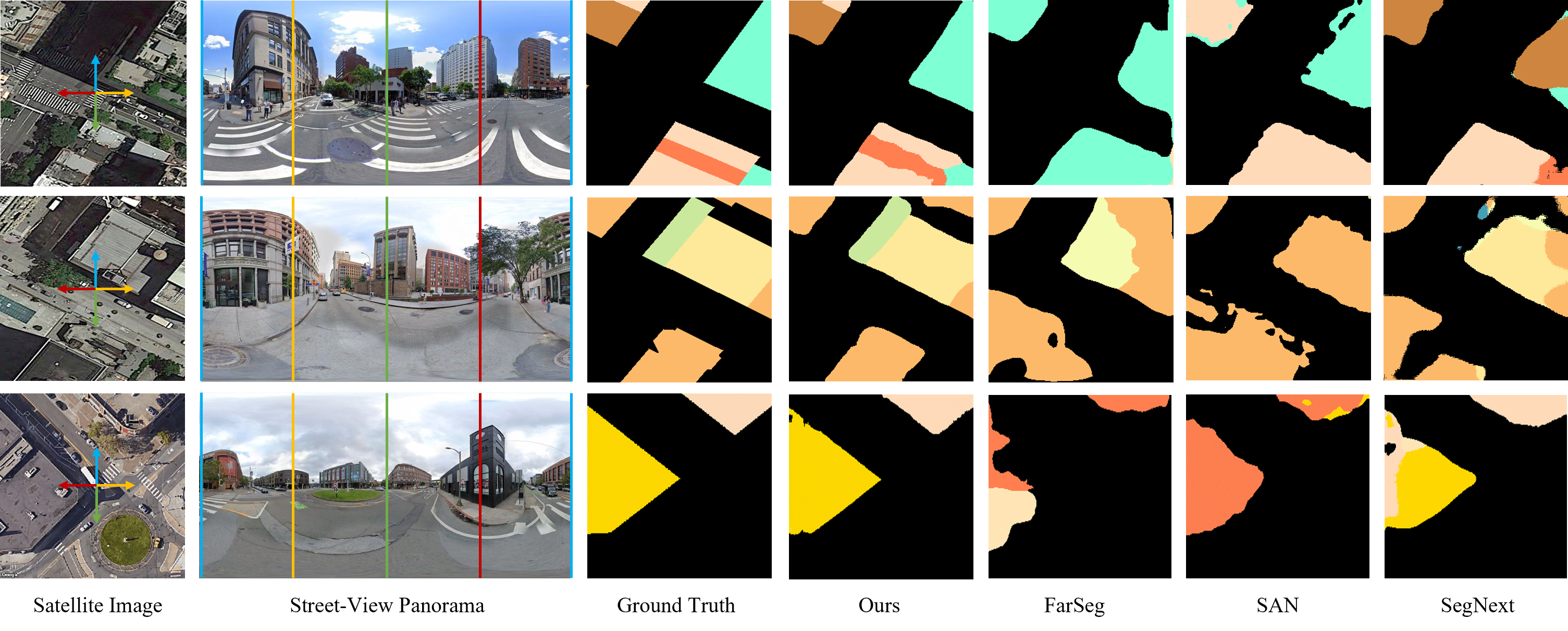}
\small 
\caption{\textbf{Comparisons of SG-BEV (Ours) and Satellite-Based Methods for Fine-Grained Segmentation.} The first two rows show results of OmniCity on land use (first row) and floor level (second row) segmentation tasks. The third row presents land use predictions of Vigor. The street-view panoramas, from left to right, correspond to a 360-degree clockwise rotation starting from the north direction in the satellite imagery.}
\label{fig:seg_result}
\end{figure*}

\begin{table*}
  \centering
  \caption{Comparison with satellited-based semantic segmentation methods on different datasets, in terms of mIoU and Acc metrics (\%).}
  \resizebox{0.95\textwidth}{!}{
  \begin{tabular}{@{}lccccccccccccc@{}}
    \toprule
    \multirow{3}{*}{Method} & \multicolumn{4}{c}{OmniCity} & \multicolumn{4}{c}{Brooklyn} & \multicolumn{2}{c}{Boston}  & \multicolumn{2}{c}{Vigor}\\
    \cmidrule(r){2-5} \cmidrule(l){6-9} \cmidrule(l){10-11} \cmidrule(l){12-13}
    & \multicolumn{2}{c}{Land use} & \multicolumn{2}{c}{Floor} & \multicolumn{2}{c}{Land use} & \multicolumn{2}{c}{Floor} & \multicolumn{2}{c}{Land use}  & \multicolumn{2}{c}{Land use}\\
    \cmidrule(r){2-3} \cmidrule(l){4-5} \cmidrule(l){6-7} \cmidrule(l){8-9} \cmidrule(l){10-11}\cmidrule(l){12-13}
     & mIOU & Acc & mIOU & Acc & mIOU & Acc & mIOU & Acc & mIOU & Acc & mIOU & Acc\\
    \midrule
    FarSeg \cite{zheng2020foreground} & 26.27 & 68.62 & 17.42 & 71.75 & 31.71 & \textbf{80.08} & 26.93 & 72.31 & 28.62 & 72.08 & 28.49 & 74.49\\
    SAN \cite{xu2023side} & 24.49 & 65.48 & 19.89 & 70.66 & 31.83 & 75.41 & 25.39 & 69.69 & 29.05 & \textbf{77.63} & 29.03 & 72.95\\
    SegNext \cite{guo2022segnext} & 31.38 & 70.31 & 25.27 & 72.27 & 36.85 & 76.68 & 34.55 & 76.01 & 32.55 & 76.81  & 34.92 & 76.13\\
    Ours & \textbf{37.54} & \textbf{76.13} & \textbf{40.64} & \textbf{77.82} & \textbf{47.19} & 78.43 & \textbf{49.51} & \textbf{79.00} & \textbf{39.72} & 77.39 & \textbf{41.70} & \textbf{76.81}\\
    \bottomrule
  \end{tabular}
  }
  \vspace{-3pt}
  \label{tab:seg_res_table}
\end{table*}

\begin{figure*}[htbp]
\centering
\includegraphics[width=\linewidth]{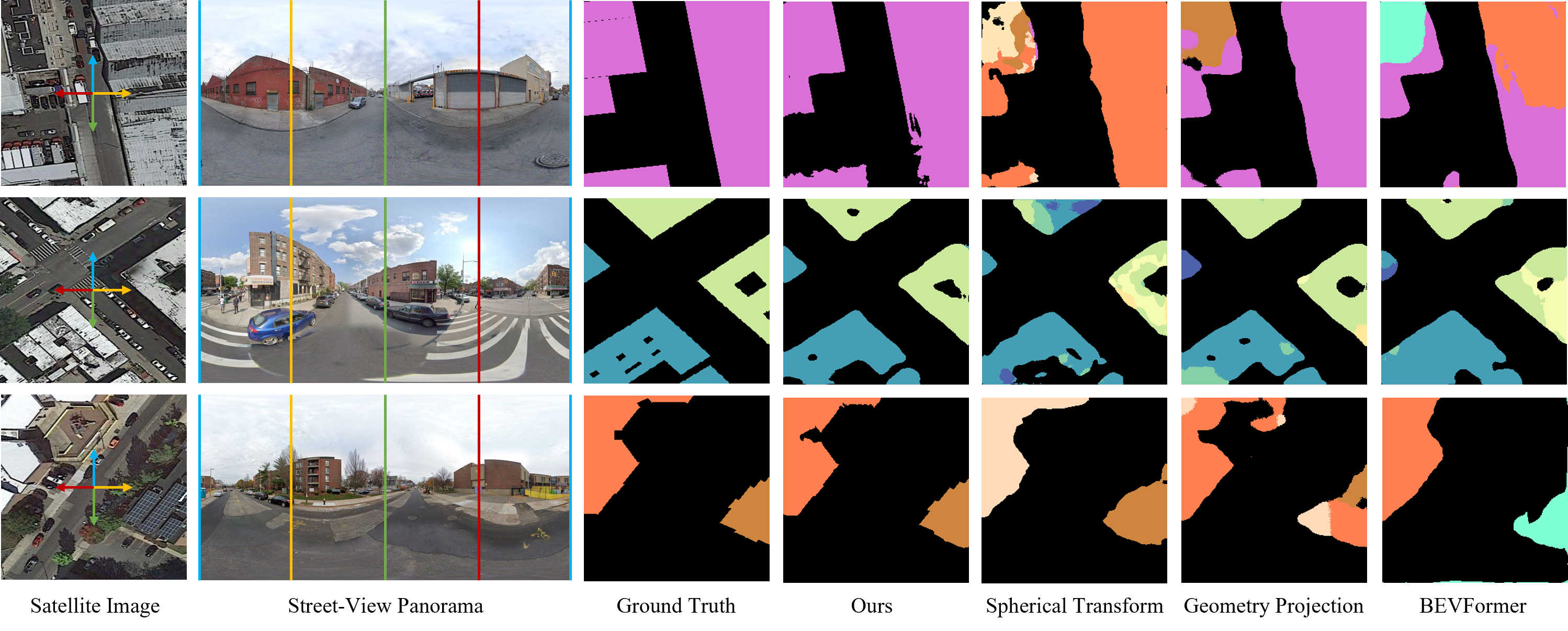}
\small 
\caption{\textbf{Comparisons of SG-BEV (Ours) and Other Cross-View Methods for Fine-Grained Segmentation.} The first two rows display results of Brooklyn on land use (first row) and floor level (second row) segmentation tasks. The bottom row illustrates land use segmentation predictions of Boston. The street-view panoramas, from left to right, correspond to a 360-degree clockwise rotation starting from the north direction in the satellite imagery.}
\label{fig:bev_result}
\end{figure*}

\begin{table*}
  \centering
  \caption{Comparison with cross-view transformation methods on different datasets, in terms of mIoU and Acc metrics (\%).}
  \resizebox{0.95\textwidth}{!}{
  \begin{tabular}{@{}lcccccccccccc@{}}
    \toprule
    \multirow{3}{*}{Method} & \multicolumn{4}{c}{OmniCity} & \multicolumn{4}{c}{Brooklyn} & \multicolumn{2}{c}{Boston }  & \multicolumn{2}{c}{Vigor}\\
    \cmidrule(r){2-5} \cmidrule(l){6-9} \cmidrule(l){10-11} \cmidrule(l){12-13}
    & \multicolumn{2}{c}{Land use} & \multicolumn{2}{c}{Floor} & \multicolumn{2}{c}{Land use} & \multicolumn{2}{c}{Floor} & \multicolumn{2}{c}{Land use}  & \multicolumn{2}{c}{Land use}\\
    \cmidrule(r){2-3} \cmidrule(l){4-5} \cmidrule(l){6-7} \cmidrule(l){8-9} \cmidrule(l){10-11}\cmidrule(l){12-13}
     & mIOU & Acc & mIOU & Acc & mIOU & Acc & mIOU & Acc & mIOU & Acc & mIOU & Acc\\
    \midrule
    ST \cite{wang2023fine} & 27.07 & 70.06 & 18.06 & 69.11 & 37.65 & 79.11 & 32.72 & 75.23 & 27.02 & 76.95 & - & -\\
    GP \cite{shi2023boosting} & 33.28 & 71.98 & 27.66 & 74.91 & 39.70 & \textbf{79.60} & 36.09 & 77.05 & 32.43 & 78.53  & - & -\\ BEVFormer \cite{li2022bevformer} & 32.17 & 71.17 & 30.81 & 75.88 & 42.96 & 78.32 & 44.59 & 76.11 & 37.16 & \textbf{78.63} & 37.33 & 76.23\\
    Ours & \textbf{37.54} & \textbf{76.13} & \textbf{40.64} & \textbf{77.82} & \textbf{47.19} & 78.43 & \textbf{49.51} & \textbf{79.00} & \textbf{39.72} & 77.39 & \textbf{41.70} & \textbf{76.81}\\
    \bottomrule
  \end{tabular}
  }
  \vspace{-10pt}
  \label{tab:cross_view_table}
\end{table*}

\subsection{Datasets}
\textbf{OmniCity Dataset (OmniCity)} \cite{Li_2023_CVPR} encompasses street-level and satellite imagery from the Manhattan area in New York, with each street-view precisely corresponding to its satellite counterpart. This dataset provides detailed urban attributes such as land use and floor level, tailored for fine-grained cross-view semantic segmentation tasks.

\noindent\textbf{Expanded Vigor Dataset (Vigor)} \cite{zhu2021vigor}  includes the nearest street-satellite image pairs centered on satellite imagery, which are selected from the original dataset and supplemented with land use information provided by DataSF\footnote{\url{https://data.sfgov.org/Housing-and-Buildings/Land-Use/us3s-fp9q}} in the San Francisco area, to extend its application from geographic localization to cross-view semantic segmentation. As street-view and satellite images are not center-aligned, the provided offset values will be used for subsequent BEV feature translation to move them to the appropriate position.

\noindent\textbf{Brooklyn-Manhattan Dataset (Brooklyn)} combines street and satellite imagery from Brooklyn and Manhattan, utilizing land use and floor level information provided by PLUTO\footnote{\url{https://www.nyc.gov/site/planning/data-maps/open-data/dwn-pluto-mappluto.page}}. Building upon the OmniCity, we have optimized the data collection step size to reduce overlap in satellite imagery, expanded the coverage of dataset, and selected higher-quality street-view images.

\noindent\textbf{Boston Dataset (Boston)} contains street-satellite image pairs across the entire city of Boston, supplemented with land use information provided by Boston Maps\footnote{\url{https://data.boston.gov/dataset/boston-buildings-with-roof-breaks}}. The dataset maintains the same rigorous selection and division criteria as the Brooklyn.

To comprehensively evaluate the robustness and generalization capabilities of models, all datasets are strictly divided according to geographic regions, with a 4:1 training to test set ratio. We have provided a comparison of different data partition methods in the supplementary materials.

\subsection{Experimental Settings}


\hspace*{1em}\textbf{Evaluation Metrics.} Our study utilizes mean intersection over union (mIOU) and overall accuracy (Acc) to evaluate our cross-view semantic segmentation task. mIOU measures the overlap between the model's predicted regions and the true regions, while Acc assesses the proportion of samples correctly predicted by the model. In our task, it involves perceptive segmentation of two fine-grained attributes of buildings: land use and the number of floors.

\textbf{Comparison Methods.} Our comparison methods are divided into two categories: The first category involves segmenting satellite imagery using state-of-the-art methods, including FarSeg \cite{zheng2020foreground}, SAN \cite{xu2023side} and SegNext \cite{guo2022segnext}. The second category comprises cross-view methods that transform street view imagery into a satellite view and then combine it with satellite features, specifically Spherical Transform (ST) \cite{wang2023fine}, Geometric Projection (GP) \cite{shi2023boosting} and BEVFormer \cite{li2022bevformer}. For these three cross-view methods, we consistently use the same satellite feature extraction and fusion module as in our approach. In our comparisons, we do not apply ST and GP to the Vigor dataset because these methods require the use of centrally aligned street view images for projection, which is not the case with the Vigor dataset. Furthermore due to the lack of temporal information in our data, we retain the \(Z\)-direction expansion capability in BEVFormer but only utilize its spatial attention module. All comparison methods follow the best settings.
%



\textbf{Implementation Details.} Our network employs SegNext with it variant MSCAN-B2 \cite{guo2022segnext} with pre-trained weights on Cityscapes \cite{cordts2016cityscapes} as the feature extractors for street-view and satellite imagery utilizing non-share weights. Satellite images and the BEV transformations derived from ground images are uniformly sized at 256 × 256 across all datasets, corresponding to a sensing range of approximately 70 × 70 meters. Ours models are trained on Nvidia GeForce RTX 3090 GPUs, starting with an initial learning rate of $6e^{-5}$, which is adjusted according to a step strategy over 50 epochs. We use AdamW as the optimizer, with an epsilon of $1e^{-8}$, a weight decay of 0.01. Information about the depth estimation method used in the paper can be found in the supplementary material. More information on depth estimation and BEV dimension settings can be found in the supplementary materials.



\subsection{Performance Comparison}

\begin{table*}[t]
\centering
\begin{minipage}[t]{\columnwidth}
\centering
\caption{\textbf{Ablation study on Satellite-Guided Reprojection module, in terms of mIOU metrics (\%).}}
\begin{tabular}{@{}l@{\hspace{2pt}}c@{\hspace{2pt}}c@{\hspace{2pt}}c@{\hspace{2pt}}c@{\hspace{2pt}}c@{\hspace{2pt}}c@{}}
\toprule
\multirow{2}{*}{Method} & \multicolumn{2}{c}{OmniCity} & \multicolumn{2}{c}{Brooklyn} \\
\cmidrule(lr){2-3} \cmidrule(lr){4-5}
& Land use & Floor & Land use & Floor \\
\midrule
UNet \cite{ronneberger2015u} 
& 27.07 & 34.13 & 37.04 & 39.61 \\
UNet \cite{ronneberger2015u} + DGR  & 32.52 & 37.01 & 42.33 & 45.06 \\
UNet \cite{ronneberger2015u}+ SGR & \textbf{36.67} & \textbf{40.53} & \textbf{47.10} & \textbf{48.55} \\
\midrule
SegNext \cite{guo2022segnext}
& 32.10 & 30.56 & 42.43 & 43.55 \\
SegNext \cite{guo2022segnext} + DGR & 35.49 & 37.64 & 45.16 & 48.17 \\
SegNext \cite{guo2022segnext} + SGR& \textbf{37.54} & \textbf{40.64} & \textbf{47.19} & \textbf{49.51} \\
\bottomrule
\end{tabular}
\par
\raggedright
\footnotesize{\hspace{15pt}DGR: Depth-Guided reprojection, utilizes only depth. \\
\hspace{15pt}SGR: Satellite-Guided reprojection, utilizes both satellite and depth.}
\label{tab:Ablation1}
\end{minipage}
\hfill
\begin{minipage}[t]{\columnwidth}
\centering
\caption{\textbf{Ablation Study on Cross-View Feature Fusion module, in terms of mIOU metrics (\%).}}
\begin{tabular}{@{}l@{\hspace{2pt}}c@{\hspace{2pt}}c@{\hspace{2pt}}c@{\hspace{2pt}}c@{\hspace{2pt}}c@{\hspace{2pt}}c@{}} 
\toprule
\multirow{2}{*}{Method} & \multicolumn{2}{c}{OmniCity} & \multicolumn{2}{c}{Brooklyn} \\
\cmidrule(l){2-3} \cmidrule(l){4-5}
& Land use & Floor & Land use & Floor \\
\midrule
ConcatFusion & 31.78 & 29.34 & 42.12 & 42.28 \\
Cross-View Fusion & \textbf{32.10} & \textbf{30.56} & \textbf{42.43} & \textbf{43.55} \\
\midrule
DGR + ConcatFusion  & 33.85 & 35.13 & 43.08 & 46.19 \\
DGR + Cross-View Fusion & \textbf{35.49} & \textbf{37.64} & \textbf{45.16} & \textbf{48.17} \\
\midrule
SGR + ConcatFusion & 36.74 & 39.69 & 46.73 & 49.28 \\
SGR + Cross-View Fusion & \textbf{37.54} & \textbf{40.64} & \textbf{47.19} & \textbf{49.51} \\
\bottomrule
\end{tabular}
\par
\raggedright
\label{tab:Ablation2}
\end{minipage}
\vspace{-10pt}
\end{table*}

\hspace*{1em}
\textbf{Compare to Satellite-Based methods.} As shown in Table \ref{tab:seg_res_table}, our cross-view segmentation results demonstrate significant improvements over segmentation using satellite imagery alone. In experiments across four datasets, our method showed an increase in mIOU for building category prediction and floor count prediction tasks by 7.61\% and 15.16\%, respectively, compared to the best satellite-based segmentation method. Predicting floor counts from a satellite view is more challenging than classifying building types, as floor count information is primarily concentrated on the facades. Our approach is more effective in this task, highlighting the efficacy of integrating street-view data for fine-grained building attribute segmentation tasks. As observed in Figure \ref{fig:seg_result}, other methods using only satellite imagery could delineate building outlines but struggled with fine-grained attribute perception of buildings. Our method achieves multi-perspective perception of building attributes, not only effectively segmenting building contours but also distinguishing between different building attributes. 
\textbf{Compare to Cross-View methods.} As shown in Table \ref{tab:cross_view_table}, we compare our method with other Cross-View methods. Compared to geometric projection methods, our approach showed an average increase in mIOU of 6.35\% and 13.20\%, respectively. As evident from Figure \ref{fig:bev_result}, the two cross-view geometric projection methods maintain good geometric fidelity only near the camera, showing significant distortion in farther areas, leading to loss of street-view feature projection and reduced segmentation performance. Among the methods implemented through geometric projection, Geometry Projection (GP) performed better than Spherical Transform (ST), as it directly projects features instead of converting them into image projections.

Additionally, compared to BEVFormer, our method, with the addition of the SGR module, demonstrates a significant improvement, with an average increase in mIOU of 4.13\% and 7.38\%. From Figure \ref{fig:bev_result}, it is observed that BEVFormer approach may correctly identify the category at the building edges, but not accurately inside the building. This is linked to BEVFormer's limitation in effectively projecting features only to building edges. More visualization results will be shown in the supplementary materials.


\subsection{Ablation study}

\hspace*{1em}
\textbf{Satellite-Guided Reprojection Performance.}
In our ablation experiments, we employed UNet \cite{ronneberger2015u} and SegNext \cite{guo2022segnext} as image encoders to validate the contribution of our proposed Satellite-Guided Reprojection module (SGR). And in order to more fully verify the role of our SGR, we also extract the Depth-Guided Reprojection (DGR) module in SGR separately for experiments. When using the DGR module, \(\alpha\) was set to a fixed value, and satellite footprint range restrictions were not applied. We will compare our method SGR with two different approaches: directly using BEV feature projection, and the second using DGR in the BEV feature projection, while maintaining the same subsequent feature fusion steps. As shown in Table \ref{tab:Ablation1}, with the DGR module, the mIOU for U-Net improved by 4.77\%, and for SegNext by 4.46\%. This shows that using depth information effectively disperses street-view features, concentrated on building edges, throughout the building area. The addition of the Satellite-Guidance further increased the mIOU for U-Net by 3.98\% and for SegNext by 2.40\%, demonstrate the importance of the supporting role of satellite information. Additionally, the performance of our SGR module was significantly enhanced in both SegNext and UNet architectures, convincingly demonstrating the SGR module's outstanding role in optimizing BEV feature distribution, resolving the issue of concentrated building edge features, and substantially improving task performance.



\textbf{Cross-View Feature Fusion.} 
To demonstrate the effectiveness of our proposed fusion strategy, we conducted a series of ablation experiments on the OmniCity and Brooklyn datasets. We will use a feature fusion method that directly concatenates satellite and BEV features along the channel dimension (ConcatFusion) as the baseline for comparison. Additionally, we further explore the impact of the reprojection module on the feature fusion module. We compared the performance of the Cross-View feature fusion module without using any reprojection module, and when using DGR and SGR modules. As shown in Table \ref{tab:Ablation2}, the Cross-View feature fusion module improved the performance of both tasks, with the mIOU averaging 0.78\%, 2.05\% and 0.61\% improvements respectively. We found that the most significant improvements occurred when using the DGR module. This is because the offsets generated by the DGR module can be unstable, potentially causing features to exceed the boundaries of buildings, leading to misalignment. Our feature fusion module includes an alignment process that effectively mitigates this issue. However, as SGR is guided by satellite information, the misalignment between different features is less pronounced, leading to a reduced performance improvement from Cross-View fusion. These results validate the effectiveness of our feature fusion module, with more efficient feature integration enhancing task performance under various conditions.

%% file: sec/5_conclusion.tex
\section{Conclusion}
\label{sec:conclusion}
\hspace*{1em}In this paper, we proposed SG-BEV, a novel satellite-guided BEV fusion method for cross-view semantic segmentation, focusing on the fine-grained attributes of buildings. Utilizing BEV method coupled with our proposed Satellite-Guided Reprojection module, our method precisely converts features from the street view to satellite view, subsequently merging them with satellite imagery features, producing fine-grained building attribute segmentation results. Our SG-BEV demonstrates significant performance improvements compared to state-of-the-art satellite-based methods and cross-view methods, with an mIOU increase over 10.13\% and 5.21\% averaged on four datasets. SG-BEV represents a novel attempt at cross-view semantic fusion, achieving a comprehensive understanding of buildings from different perspectives. We hope that our work will inspire further research into cross-view semantic segmentation tasks.

\begin{flushleft}
\textbf{Acknowledgements.} This project was funded in part by National Natural Science Foundation of China (Grant No. 42201358) and Shanghai Artificial Intelligence Laboratory.
\end{flushleft}

%% file: sec/Reference.tex
{
    \small
    \bibliographystyle{plain}
    \bibliography{references}
}

%% file: sec/X_suppl.tex
\clearpage
\maketitlesupplementary

In this supplementary material, we provide additional details about our method in Section \ref{sec:method}, a detailed introduction to the datasets in Section \ref{sec:datasets}, specifics of experimental settings and additional experimental results in Section \ref{sec:experient}.

\appendix
\section{Additional details of our proposed method}
\label{sec:method}
\subsection{Visualization of point clouds reprojection using our SGR module}

\begin{figure}[ht]
\centering
\includegraphics[width=0.9\linewidth]{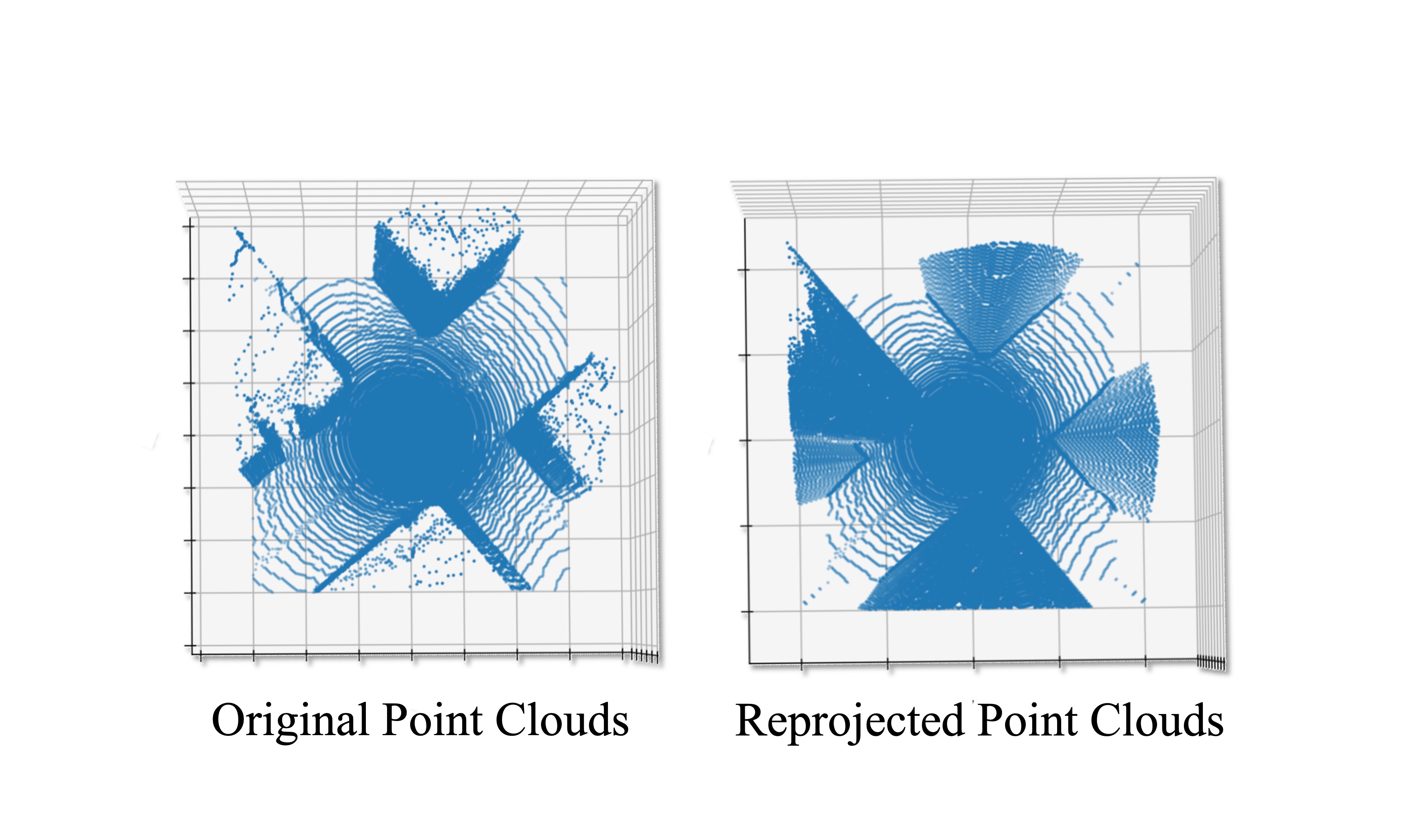}
\caption{Distribution of point clouds obtained before and after using SGR module.}
\label{fig:pointcloud}
\end{figure}

\hspace*{1em}In this section, we provide additional visualizations of point clouds offset obtained before and after using Satellite-Guided Reprojection (SGR) module (introduced in Section 3 of the main paper). 
The original point clouds obtained through depth information projection often appear as shown on the left side of Figure \ref{fig:pointcloud}, where the point clouds are noticeably concentrated along the edges of building exteriors, leaving larger interior areas of the buildings devoid of point clouds distribution. After passing through the SGR module, the reprojected point clouds gradually shift from the exterior walls of the buildings to their interiors, as depicted on the right side of Figure \ref{fig:pointcloud}.





\subsection{Visualization comparison of different cross-view projection algorithms}



\begin{figure}[ht]
\centering
\includegraphics[width=\linewidth]{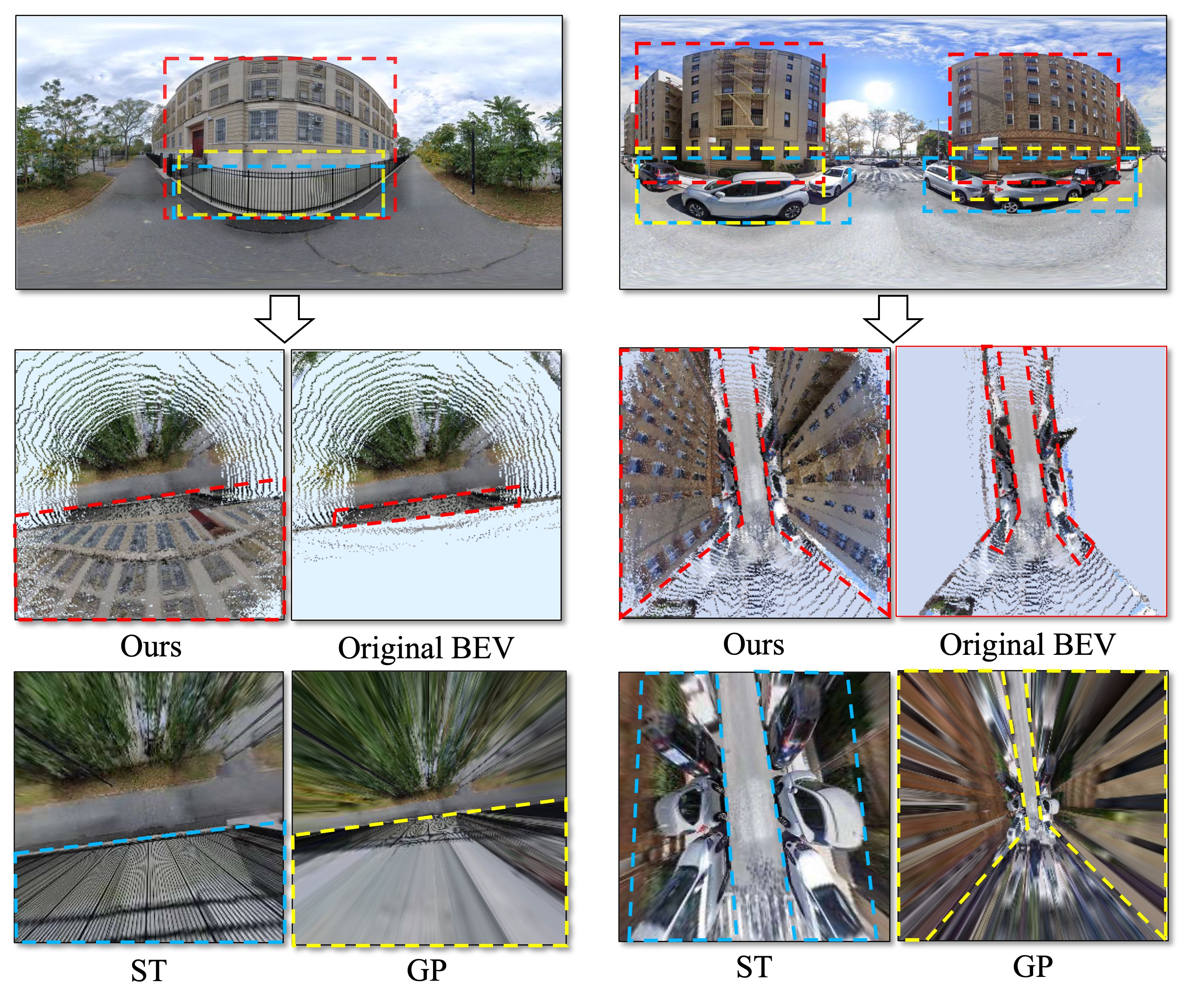}
\caption{Intuitive visualization comparisons of different cross-view projection algorithms.}
\label{fig:bev}
\end{figure}

\hspace*{1em}To intuitively demonstrate the feature mapping results of various methods, RGB values are utilized to substitute the features requiring mapping, facilitating a visual comparison. 
As clearly shown in Figure \ref{fig:bev}, the geometric projection methods Spherical Transform (ST) \cite{wang2023fine} and Geometric Projection (GP) \cite{shi2023boosting} can only map features of low building facades, such as railings and low walls, and these features are severely distorted. The original BEV method leads to a concentration of features at the wall locations, resulting in sparse interior building features. Our approach employs the SGR module to efficiently map facade features onto the BEV plane, ensuring maximal transfer while maintaining visual continuity.

\subsection{More details about the structure of cross-view feature fusion module}

\begin{figure}[ht]
\centering
\includegraphics[width=\linewidth]{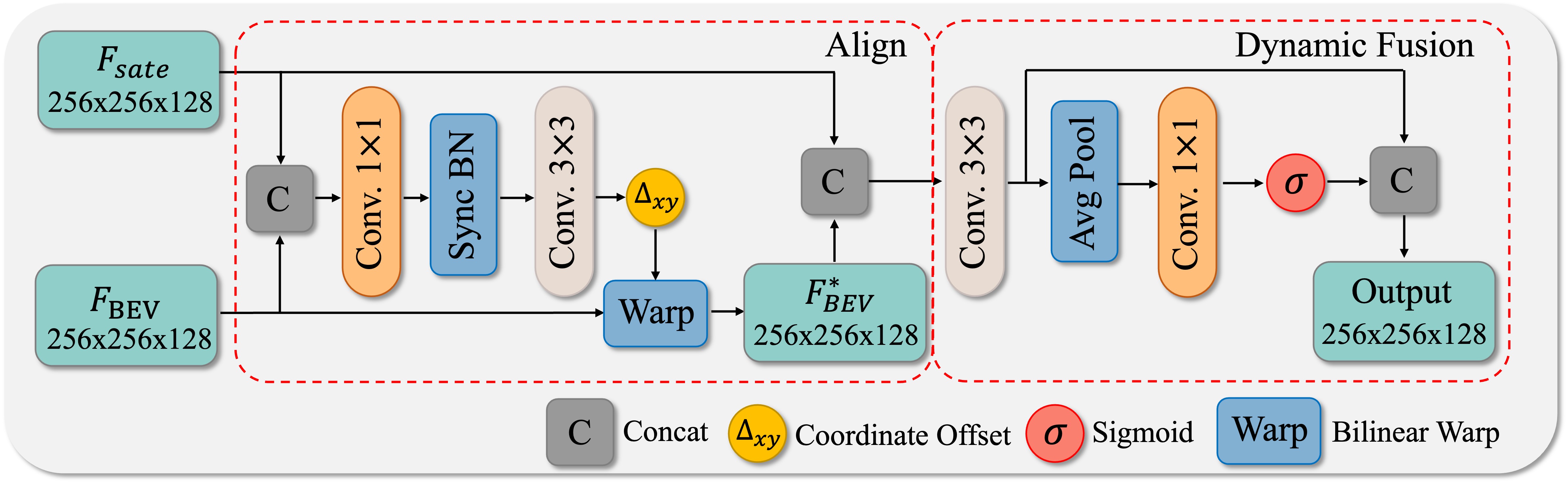}
\caption{The structure of our cross-view feature fusion module.}
\label{fig:fusion}
\end{figure}

\hspace*{1em}The input satellite features and Bird's Eye View (BEV) features are both unified within the same top-down perspective feature space. As shown in Figure \ref{fig:fusion}, we initially employ an align module to shift the BEV features for alignment with the satellite features. 
Subsequently, a dynamic fusion module is employed to optimize the feature fusion process.

\section{Additional details of the datasets}
\label{sec:datasets}

\subsection{Details of panorama depth maps}

\begin{figure}[ht]
\centering
\includegraphics[width=\linewidth]{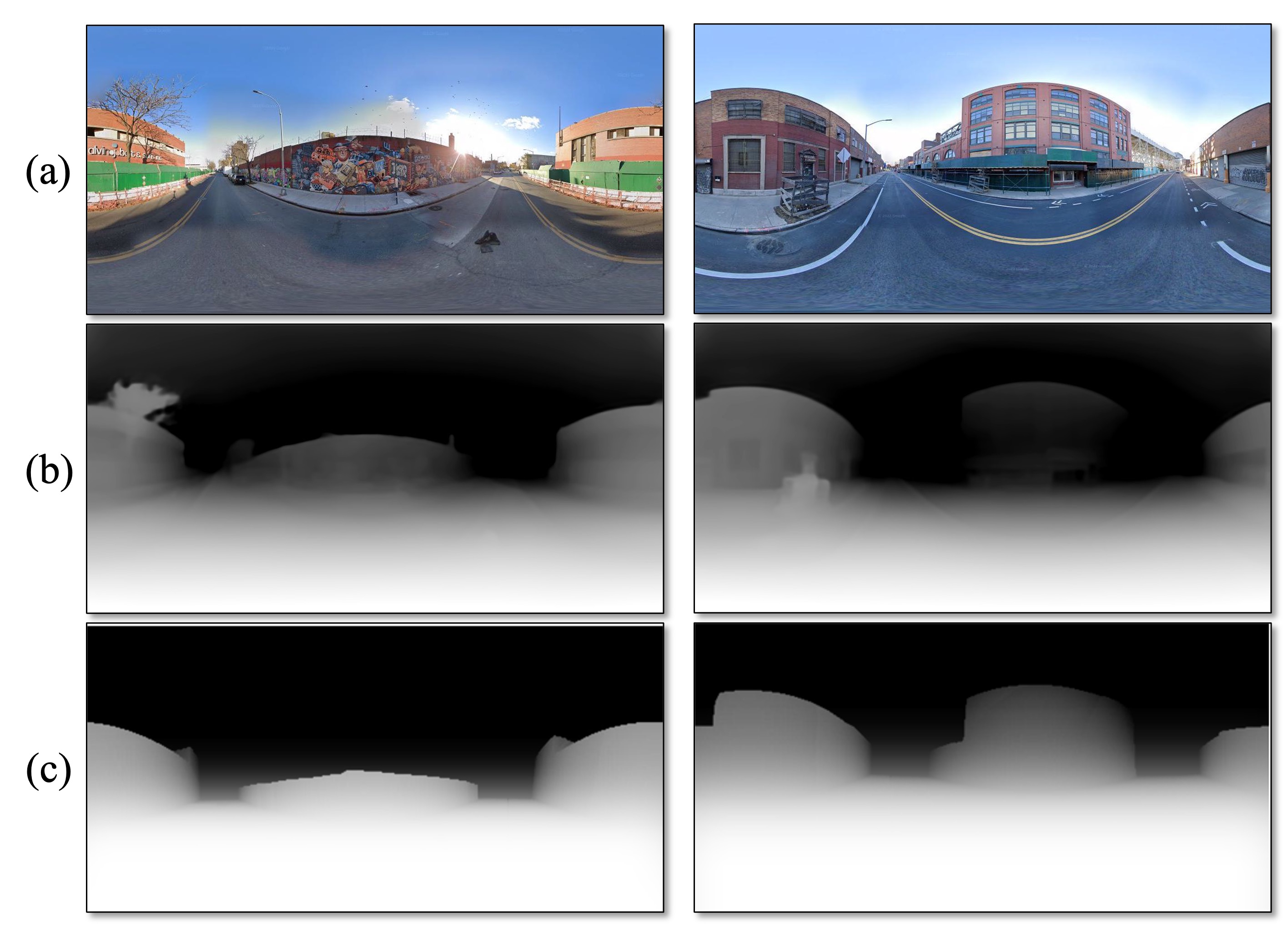}
\caption{Depth estimation result based on street-view image. (a) Street-view image from OmniCity dataset. (b) Depth estimation results from ZoeDepth \cite{bhat2023zoedepth}. (c) Depth maps provided by Google Street View Download 360.}
\label{fig:depth}
\end{figure}

\hspace*{1em}For monocular depth estimation of street-view panorama images, we can employ established depth estimation algorithms like ZoeDepth \cite{bhat2023zoedepth}, or use Google Street View Download 360\footnote{\url{https://svd360.istreetview.com/}\label{360}} to download corresponding depth maps, as shown in Figure \ref{fig:depth}. Although both methods yield favorable results, the depth maps provided by Google Street View Download 360 are more accurate for depth estimation in building areas. Hence, we use it as the data source of depth estimation information in our method.

\subsection{Details of each dataset}

\hspace*{1em}For the public Vigor \cite{zhu2021vigor} dataset, we supplemented it with land use information provided by DataSF\footnote{\url{https://data.sfgov.org/Housing-and-Buildings/Land-Use/us3s-fp9q}\label{datasf}} in the San Francisco area. Additionally, to facilitate subsequent BEV tasks, we augmented each street-view image in both the Vigor and OmniCity \cite{Li_2023_CVPR} datasets with depth maps using Google Street View Download 360\footref{360}.
For our self-collected Brooklyn and Boston datasets, we used property data provided by PLUTO\footnote{\url{https://www.nyc.gov/site/planning/data-maps/open-data/dwn-pluto-mappluto.page}\label{pluto}} and Boston Maps\footnote{\url{https://data.boston.gov/dataset/boston-buildings-with-roof-breaks}\label{bostonmap}}, 
as well as the OpenStreetMap (OSM) building outlines to obtain attribute data of individual buildings following the approach used in OmniCity dataset.
Compared with OmniCity dataset, Brooklyn dataset covers the entire Brooklyn and Manhattan areas, with the step distance of street view images  increased to 97.5 meters to reduce the overlap of satellite images. The Boston dataset covers the urban area of Boston with the same step distance of street view images as Brooklyn dataset.
For each dataset, we strictly divide the training and test samples by regional partition (train: test = 4: 1). Table \ref{tab:dataset_numbers} shows the number of training and test samples of the four datasets, along with the corresponding categories for each dataset shown in Table \ref{tab:dataset_labels}.

\begin{table}[htb]
\centering
\caption{Number of training and test samples in each dataset. }
\footnotesize
\begin{tabular}{lcccc}
\toprule
Dataset & OmniCity \cite{Li_2023_CVPR} & Vigor \cite{zhu2021vigor} & Brooklyn & Boston\\
\midrule
Train & 14,400 & 11,960 & 7,600 & 7,036 \\
Test & 3,600 & 2,990 & 1,900 & 1,759 \\
\bottomrule
\end{tabular}
\label{tab:dataset_numbers}
\end{table}

\subsection{The impact of different data partition methods}


\hspace*{1em}As mentioned above, the training and test sets for all datasets were collected through regional partitioning. To further explore the impact of different data partition methods, we compare the experimental results of the OmniCity and Brooklyn datasets using random partitioning and regional partitioning. As illustrated in Table \ref{tab:diferent data splitting}, the performance metrics obtained through the random partitioning approach are remarkably higher than those achieved with regional partitioning, a trend particularly evident in the densely sampled OmniCity dataset. 
This overestimation is primarily attributed to the nature of the cross-view datasets that are densely sampled and randomly partitioned, where a single satellite image often covers multiple street view images. 
Random partitioning of training and test sets in cross-view image pairs might lead to significant overlaps among satellite images, compromising the dataset's independence and resulting in inflated performance metrics.



\begin{table}[ht]
\centering
\footnotesize
\caption{Quantitative analysis of random and regional partitioning in OmniCity and Brooklyn datasets, which indicates the overestimation of performance metrics using random partitioning.}
\begin{tabular}{l|@{\hspace{5pt}}c|@{\hspace{9pt}}c@{\hspace{9pt}}c@{\hspace{9pt}}c@{\hspace{9pt}}c}
\toprule
\multirow{3}{*}{Partition} & \multirow{3}{*}{Method} & \multicolumn{4}{c}{Dataset} \\
\cmidrule(r){3-6}
 & & \multicolumn{2}{c}{OmniCity} & \multicolumn{2}{c}{Brooklyn} \\
\cmidrule(r){3-4} \cmidrule(r){5-6}
 & & Land use & Floor & Land use & Floor \\
\midrule
\multirow{3}{*}{Random} & SegNext \cite{guo2022segnext} & 77.31 & 77.92 & 45.00 & 44.61 \\
 & BEVFormer \cite{li2022bevformer} & 79.15 & 78.86 & 48.89 & 50.32 \\
 & Ours & 80.13 & 81.88 & 52.50 & 56.09 \\
\midrule
\multirow{3}{*}{Regional} & SegNext \cite{guo2022segnext} & 31.25 & 31.19 & 35.77 & 33.16 \\
 & BEVFormer \cite{li2022bevformer} & 31.95 & 32.18 & 41.89 & 43.32 \\
 & Ours & 37.95 & 40.02 & 47.20 & 48.81 \\
\bottomrule
\end{tabular}
\label{tab:diferent data splitting}
\end{table}

\section{Additional details of experimental results}
\label{sec:experient}
\subsection{Additional experimental analysis on hyperparameter settings}

\hspace*{1em}In our Satellite-Guided Reprojection (SGR) method, the calculation of offset \(\Delta\) involved two critical hyperparameters: \(d_0\) and \(\alpha\), as shown in Eq. \ref{eq:eq3}.

\begin{equation}
\Delta = \log(1 + d - d_0) \times \alpha 
\label{eq:eq3}
\end{equation}


In our study, the parameter \(d_0\) is introduced to mitigate the feature offset from areas such as roads to building regions. Typically, \(d_0\) represents the depth of the ground within a certain range from the camera. In practical computation, if \(d\) is less than \(d_0\), the offset will not occur. 
Considering that the width of urban roads typically ranges between 10 to 14 meters\footnote{\url{https://safety.fhwa.dot.gov/geometric/pubs/mitigationstrategies/chapter3/3_lanewidth.cfm}}, 
together with the additional sidewalks of about 2 meters wide on each side,
we set \(d_0\) to approximately 10 meters. This distance represents the average distance from a vehicle to either edge of the road. With access to more specific road width information, the parameter \(d_0\) can be further optimized to achieve enhanced performance.



\begin{figure}[ht]
\centering
\includegraphics[width=\linewidth]{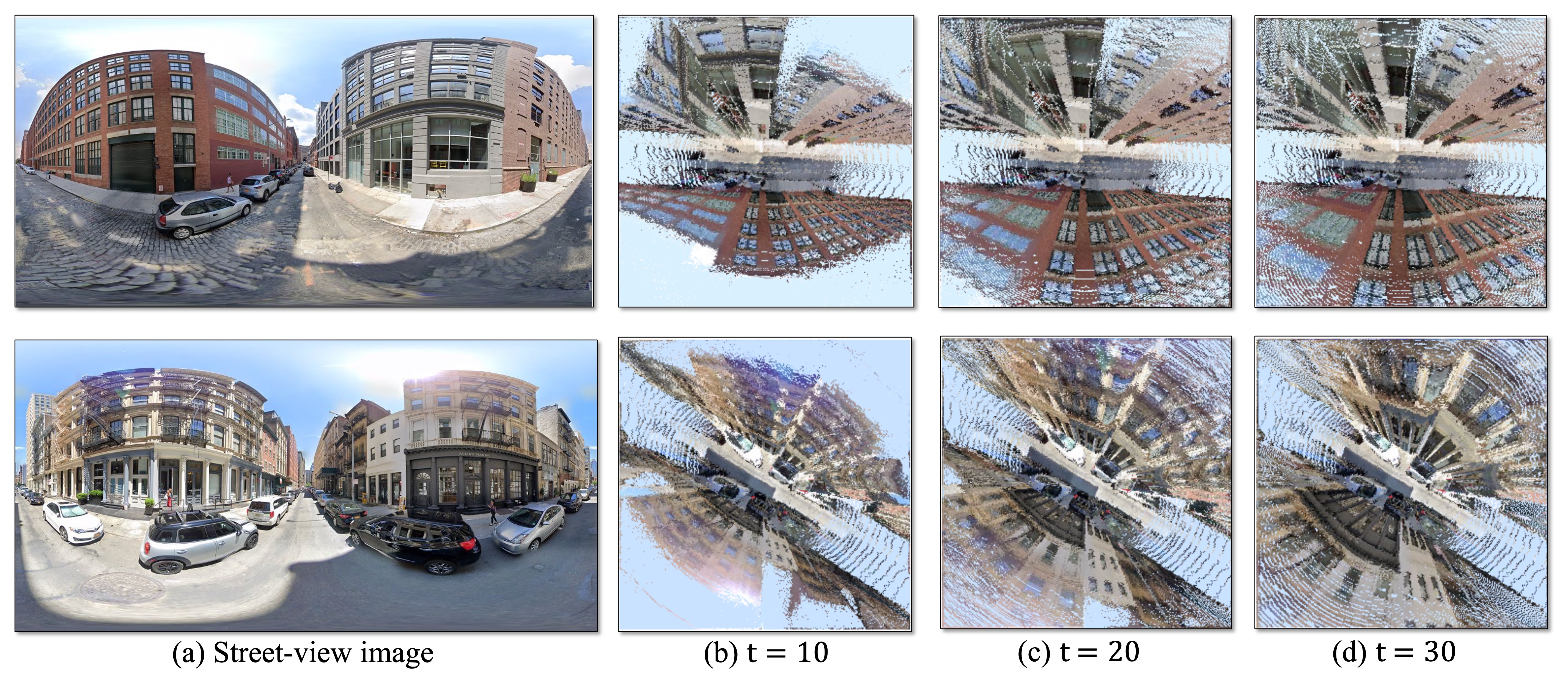} 
\caption{Visualization of the reprojection results using different $\alpha$ values.}
\label{fig:alpha}
\end{figure}

Another critical hyperparameter in our study was \(\alpha\), which determined the amplitude of variation in \(\Delta\) at different depths under the same building. Initially, the entire area is divided into \(3 \times 3\) blocks, and then the proportion of building footprint pixels in each block is calculated to obtain the pixel ratio parameter \(\rho\). Below is our formula for calculating \(\alpha\) based on \(\rho\). The hyperparameter \( t \) in the formula is adjustable. We tested three different values of \( t \): 10, 20, and 30, with a higher value of \( t \) indicating a larger amplitude of change. As shown in Table \ref{tab:alpha}, our experiments on the OmniCity and Brooklyn datasets with these \(t\) values yielded results, demonstrating that \(t = 20\) achieved the best performance. Furthermore, as illustrated in Figure \ref{fig:alpha}, \(t = 20\) is the most balanced choice in terms of visual effects. Therefore, \(t = 20\) was selected as the parameter for our experiments.



\begin{equation}
\alpha = 
\begin{cases} 
0 & \text{if } \rho \leq 0.1, \\
5 + t \times \rho & \text{if } \rho > 0.1.
\end{cases}
\label{eq:eq2}
\end{equation}



\begin{table}[ht]
\centering
\footnotesize
\caption{Quantitative analysis of different $\alpha$, in terms of mIoU (\%), which indicates that \(\alpha = 20\) yields the best performance.}
\begin{tabular}{c|cccc}
\toprule
\multirow{3}{*}{Parameter} & \multicolumn{4}{c}{Dataset} \\
\cmidrule(r){2-5}
& \multicolumn{2}{c}{OmniCity} & \multicolumn{2}{c}{Brooklyn} \\
\cmidrule(r){2-3} \cmidrule(r){4-5}
& Land use & Floor & Land use & Floor \\
\midrule
\(t = 10\) & 36.18 & 39.45 & 46.99 & 47.74 \\
\(t = 20\) & 37.54 & \textbf{40.64} & \textbf{47.19} & \textbf{49.51} \\
\(t = 30\) & \textbf{37.68} & 40.41 & 46.15 & 48.42 \\
\midrule
\end{tabular}
\vspace{-3pt}
\label{tab:alpha}
\end{table}

\subsection{Additional experimental analysis on satellite image size}

\hspace*{1em}We also provide experimental analysis to validate the model performance using satellite images of different sizes, including (1) \(128 \times 128\) pixels, (2)  \(256 \times 256\) pixels, and (3) \(512 \times 512\) pixels.

\begin{table}[ht]
\centering
\footnotesize
\caption{Quantitative analysis of different sizes of satellite images, in terms of mIoU (\%), which illustrates that \(256 \times 256\) pixels yield the best performance.}
\begin{tabular}{c|@{\hspace{3pt}}c|@{\hspace{3pt}}c@{\hspace{3pt}}c@{\hspace{3pt}}c@{\hspace{3pt}}c}
\toprule
\multirow{3}{*}{Satellite Image Size} & \multirow{3}{*}{Method} & \multicolumn{4}{c}{Dataset}\\
\cmidrule(r){3-6}
 & & \multicolumn{2}{c}{OmniCity} & \multicolumn{2}{c}{Brooklyn}\\
\cmidrule(r){3-4} \cmidrule(r){5-6}
 & & Land use & Floor & Land use & Floor\\
\midrule
\multirow{2}{*}{\(128 \times 128\)} & SegNext \cite{guo2022segnext} & 28.65 & 24.17 & 32.44 & 30.07 \\
 & SG-BEV & 36.82 & 38.23 & 46.82 & \textbf{49.89} \\
\midrule
\multirow{2}{*}{\(256 \times 256\)} & SegNext \cite{guo2022segnext} & 31.38 & 25.27 & 36.85 & 34.55 \\
 & SG-BEV & \textbf{37.54} & \textbf{40.64} & \textbf{47.19} & 49.51 \\
\midrule
\multirow{2}{*}{\(512 \times 512\)} & SegNext \cite{guo2022segnext} & 29.88 & 26.03 & 37.31 & 30.77 \\
 & SG-BEV & 32.55 & 31.98 & 43.04 & 37.35 \\
\bottomrule
\end{tabular}
\label{tab:diferent bev range}
\end{table}

\begin{figure}[ht]
\centering
\includegraphics[width=\linewidth]{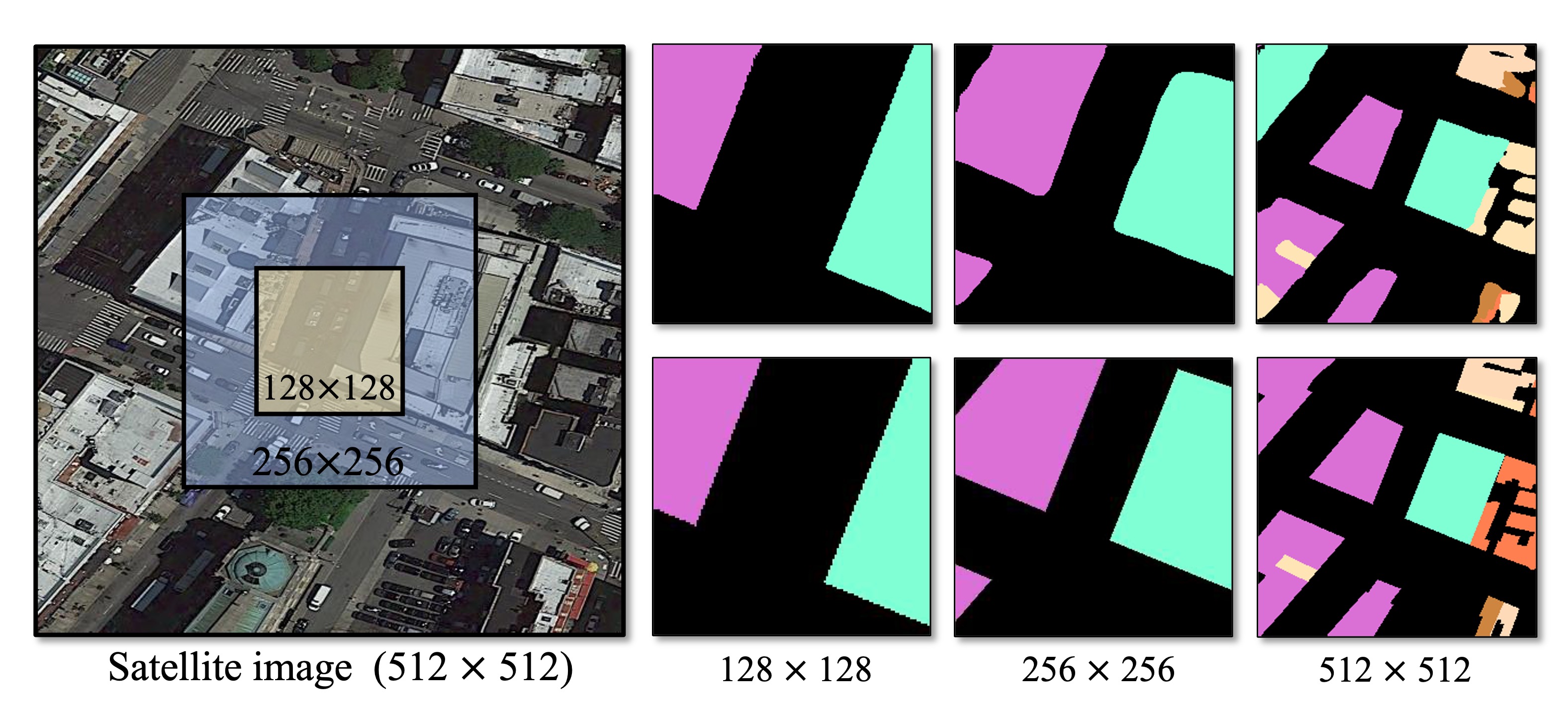}
\caption{Visualization of the semantic segmentation results by applying different sizes of satellite images. The first column shows the satellite image. The rest columns represent the semantic segmentation results obtained from satellite images of different sizes (first row) alongside their corresponding ground truths (second row).}
\label{fig:512}
\end{figure}

From Table \ref{tab:diferent bev range}, it is evident that the model performs worst when using the image size of \(512 \times 512\) pixels.
This is because with the increasing of image size, the satellite image may cover multiple blocks and streets, making it challenging for a single street-view panorama to provide sufficient effective information. 
As observed in Figure \ref{fig:512}, beyond a certain range, the model's segmentation performance begins to decline sharply. 
The model achieves better performance using image sizes of \(128 \times 128\) and \(256 \times 256\) pixels. However, too small image size may limit the model's perceptual field, which is detrimental to downstream fine-grained segmentation tasks and leads to inefficiencies in large-scale applications.
Considering the above factors, we select \(256 \times 256\) pixels as the satellite image size used in our experiments.


\subsection{More quantitative experimental results}
\hspace*{1em}The performance of our method in fine-grained attribute segmentation on four different datasets is demonstrated in Tables \ref{case2}, \ref{brooklyn}, \ref{Boston}, and \ref{Vigor}. We compare our approach with the state-of-the-art satellite-based method (SegNext \cite{guo2022segnext}) and cross-view method (BEVFormer \cite{li2022bevformer}). Our method significantly enhances the performance in almost all building categories across all datasets, demonstrating its robustness across a wide range of urban architectural styles and task attributes.


\subsection{More qualitative experimental results}

\begin{figure*}[ht]
\centering
\includegraphics[width=\linewidth]{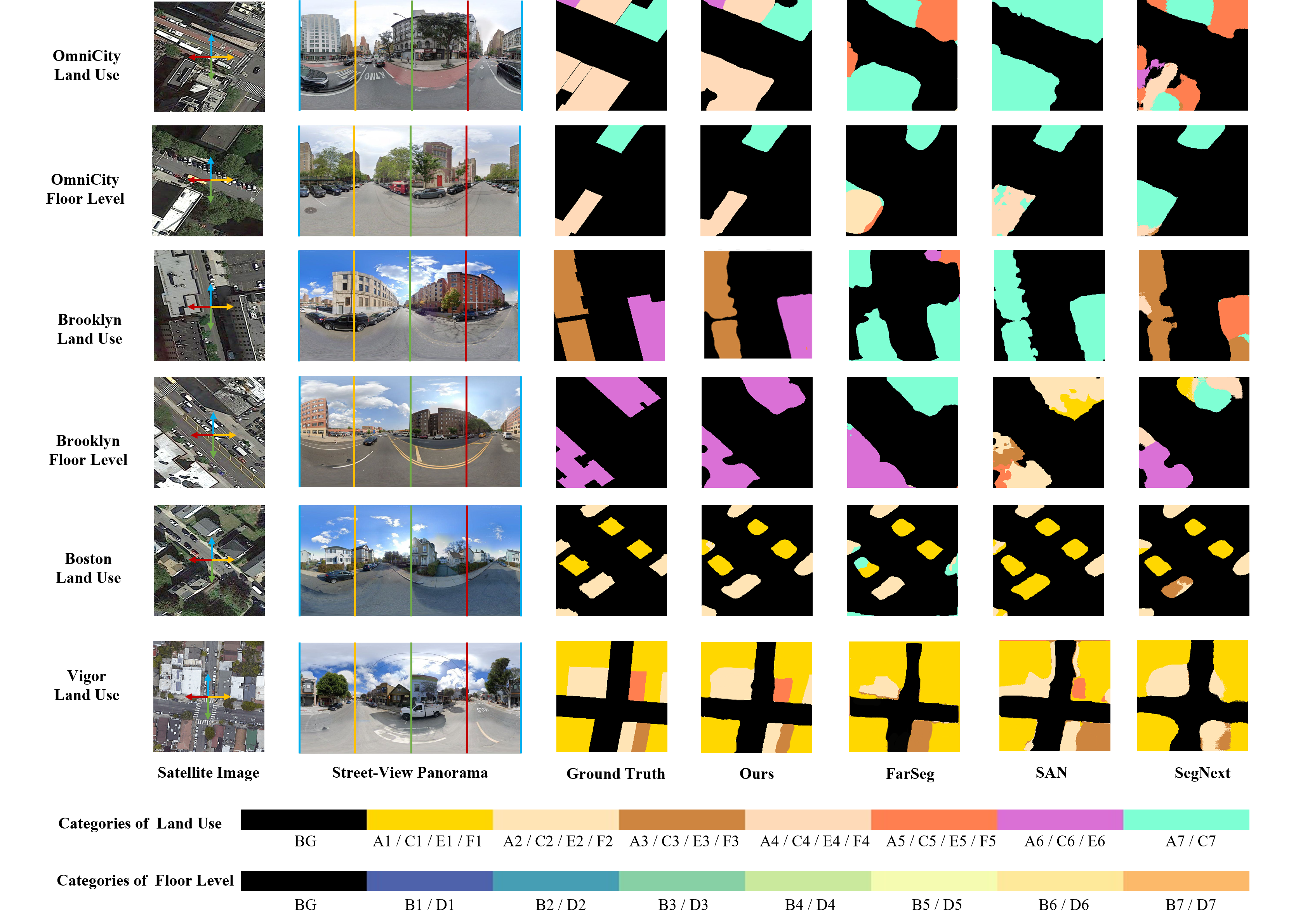}
\small 
\caption{\textbf{Comparisons of SG-BEV (Ours) and Satellite-Based Methods for Fine-Grained Segmentation.} The first two rows show results of OmniCity on land use and floor level segmentation tasks. The third and forth rows present land use and floor level segmentation results of Brooklyn. The fifth and sixth rows show the land use segmentation results of Boston and Vigor. The street-view panoramas, from left to right, correspond to a 360-degree clockwise rotation starting from the north direction in the satellite imagery.}
\label{fig:seg_result_supply1}
\end{figure*}
\begin{figure*}[ht]
\centering
\includegraphics[width=\linewidth]{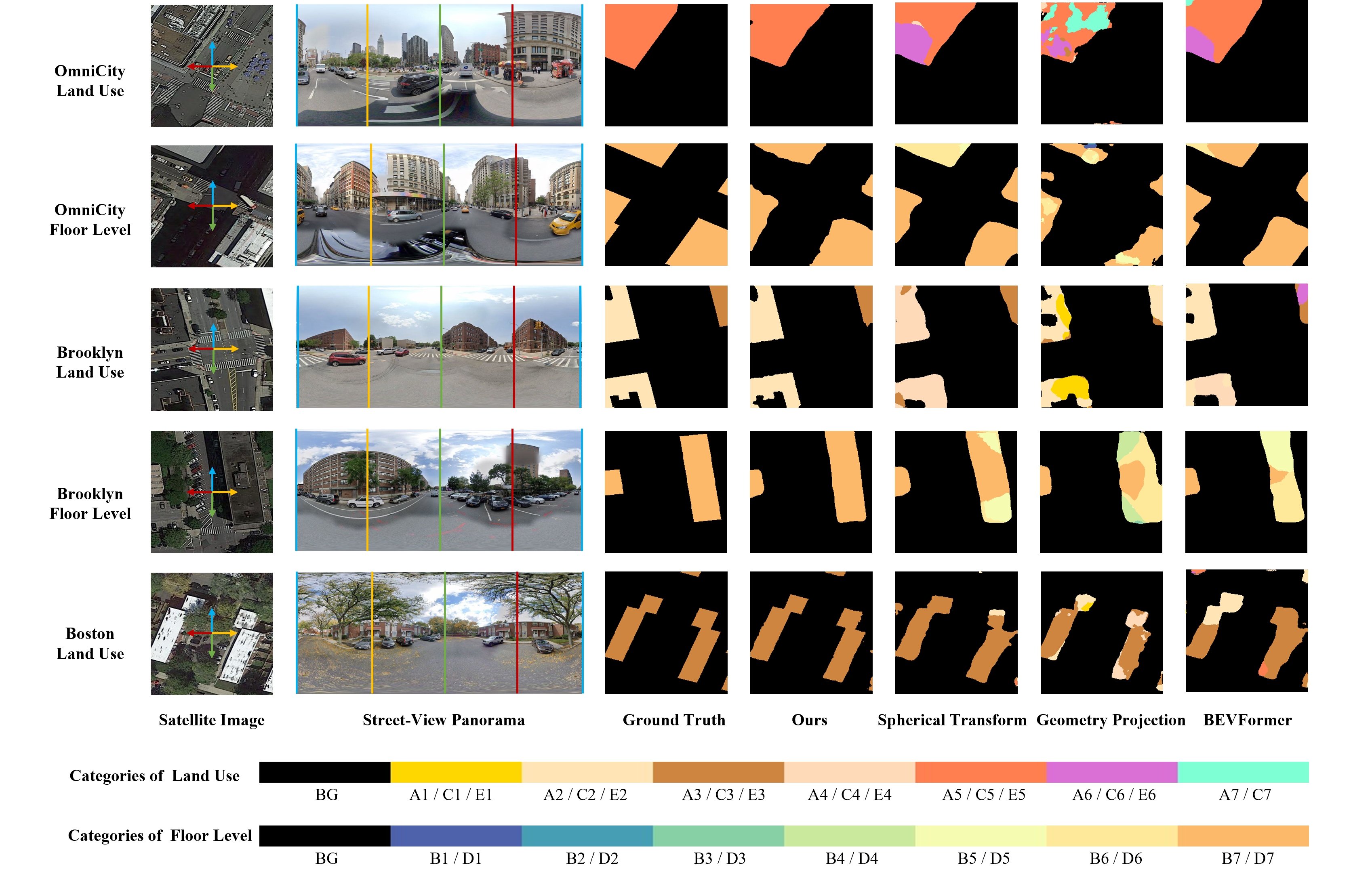}
\small 
\caption{\textbf{Comparisons of SG-BEV (Ours) and Cross-View Methods for Fine-Grained Segmentation.} The first two rows show results of OmniCity on land use and floor level segmentation tasks. The third and forth rows present land use and floor level segmentation results of Brooklyn. The fifth row shows the land use segmentation results of Boston. Results for the Vigor dataset are not included, as the offset problem in this dataset makes the Spherical Transform and Geometric Projection methods inapplicable. The street-view panoramas, from left to right, correspond to a 360-degree clockwise rotation starting from the north direction in the satellite imagery.}
\label{fig:seg_result_supply2}
\end{figure*}
\begin{figure*}[ht]
\centering
\includegraphics[width=\linewidth]{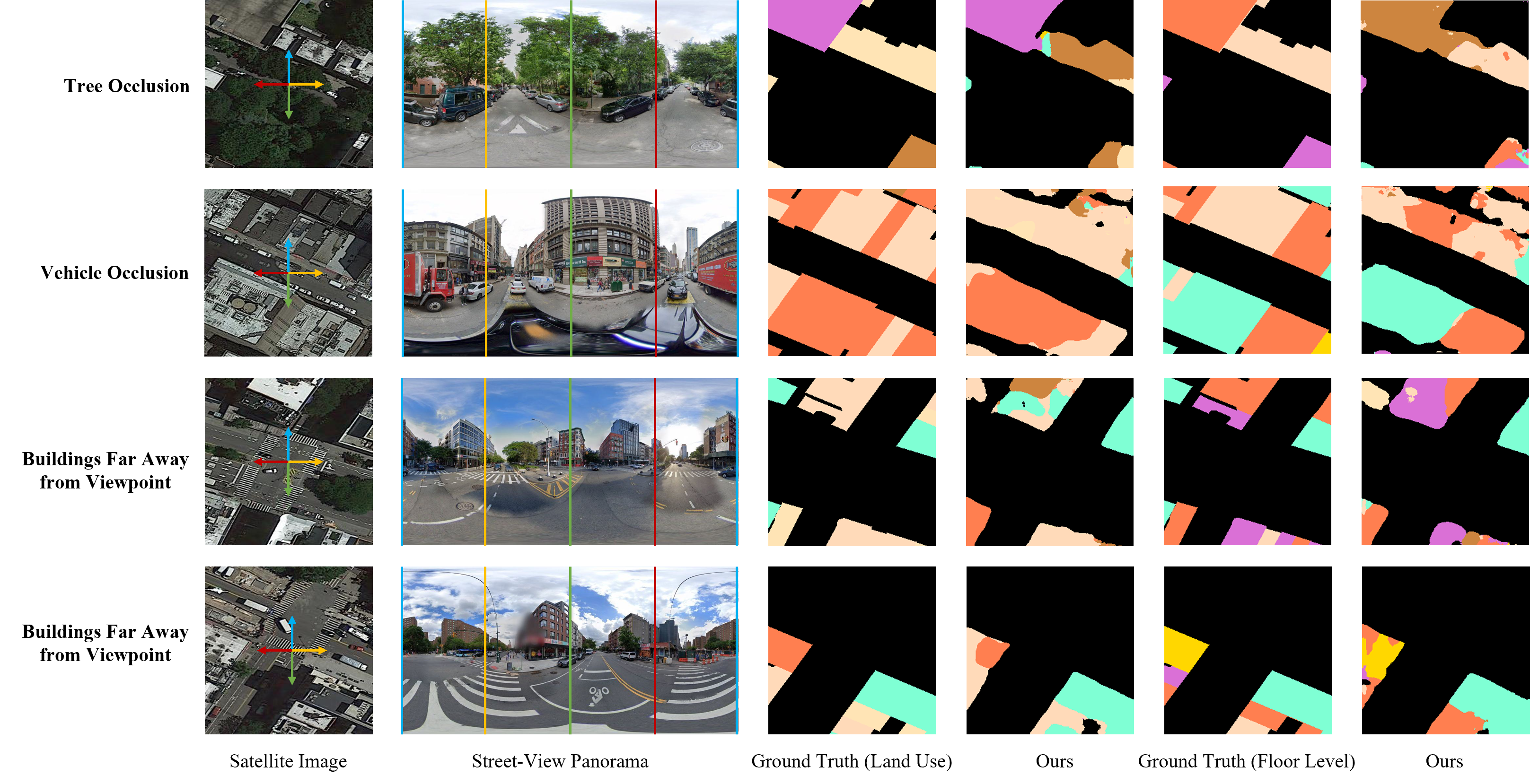}
\small 
\caption{\textbf{Limitations of the SG-BEV (Ours) Method.} The first row depicts common errors in land use and floor level segmentation tasks, primarily due to occlusion by trees. The second row presents inaccuracies in land use and floor level predictions, resulting from occlusion caused by large vehicles. The third and fourth rows display unclear segmentation results, arising from the considerable distance of buildings from the viewpoint.}
\label{fig:seg_result_supply3}
\end{figure*}

\hspace*{1em}We provide additional visualizations for various datasets. Figure \ref{fig:seg_result_supply1} illustrates the comparison of our SG-BEV method with different satellite-based method. Unlike other methods that roughly identify building outlines without discerning fine-grained attributes, our method is capable of differentiating buildings with distinct attributes. In Figure \ref{fig:seg_result_supply2}, SG-BEV is compared with various cross-view methods, demonstrating more comprehensive feature mapping within the same building, leading to more consistent internal attributes and superior performance. Moreover, Figure \ref{fig:seg_result_supply3} shows several typical failure cases, such as occlusions by trees and vehicles, or the shooting locations too far from the buildings.

\begin{table*}[ht]
\centering
\caption{Label Categories for Each Dataset. OmniCity \cite{Li_2023_CVPR} dataset contains detailed land use information. Brooklyn, Boston, and Vigor's land use information comes from PLUTO\footref{pluto}, Boston Maps\footref{bostonmap}, and DataSF\footref{datasf}, respectively.}
\label{tab:dataset_labels}
\resizebox{\textwidth}{!}{
\begin{tabular}{cccccc}
\toprule
\multirow{2}{*}{Category} & \multicolumn{2}{c}{OmniCity \cite{Li_2023_CVPR}/ Brooklyn} & Boston & Vigor \cite{zhu2021vigor} \\
\cmidrule(r){2-3} \cmidrule(l){4-5}
 & Land Use & Floor & Land Use & Land Use \\
\midrule
1 & Background (BG) & Background (BG) & Background (BG) & Background (BG) \\
2 & 1/2 Family Buildings (A1/C1) & Level 1 (B1/D1) & Industrial Manufacturing (E1) & Residential (F1) \\
3 & Walk-Up Buildings (A2/C2) & Level 2 (B2/D2) & Commercial (E2) & Mixed Use (F2) \\
4 & Elevator Buildings (A3/C3) & Level 3 (B3/D3) & High Residential (E3) & Industrial (F3) \\
5 & Mixed Residential/ Commercial (A4/C4) & Level 4 (B4/D4) & Low Medium Residential (E4) & Cultural/ Institutional/ Educational (F4) \\
6 & Office Buildings (A5/C5) & Level 5 (B5/D5) & Low Residential (E5) & Others (F5) \\
7 & Industrial/ Transportation/ Utility (A6/C6) & Level 6 (B6/D6) & Public (E6) & - \\
8 & Others (A7/C7) & Level 7 and Above (B7/D7) & - & - \\
\bottomrule
\end{tabular}
}

\end{table*}

\begin{table*}[ht]
\centering
\footnotesize
\caption{Fine-grained attribute segmentation results of different methods on the OmniCity dataset. Our method demonstrates an improvement in the mIoU by 0.95\% - 10.37\% and 0.1\% - 11.53\% for the land use attribute, respectively, and 0.57\% - 22.39\% and 0.75\% - 12.43\% for the floor level attribute, respectively, compared with the state-of-the-art.
}
\begin{tabular}{c@{\hspace{6pt}}c@{\hspace{6pt}}c@{\hspace{6pt}}c@{\hspace{6pt}}c@{\hspace{6pt}}c@{\hspace{6pt}}c@{\hspace{6pt}}c@{\hspace{6pt}}c@{\hspace{6pt}}c@{\hspace{6pt}}c@{\hspace{6pt}}c@{\hspace{6pt}}c@{\hspace{6pt}}c@{\hspace{6pt}}c@{\hspace{6pt}}c@{\hspace{6pt}}c}
\toprule
\multirow{3}{*}{Method} & \multicolumn{16}{c}{mIoU (\%) of each category} \\
\cmidrule(r){2-17}
 & \multicolumn{8}{c}{Land use} & \multicolumn{8}{c}{Floor} \\
\cmidrule(r){2-9} \cmidrule(r){10-17}
 & BG & A1 & A2 & A3 & A4 & A5 & A6 & A7 & BG & B1 & B2 & B3 & B4 & B5 & B6 & B7 \\
\midrule
SegNext \cite{guo2022segnext}& 86.31 & 13.16 & 29.53 & 19.58 & 26.70 & 34.23 & 21.75 & 19.73 & 83.17 & 7.04 & 5.69 & 7.33 & 15.17 & 16.59 & 15.41 & 51.75 \\
BEVFormer \cite{li2022bevformer}& 87.16 & 15.37 & 30.47 & 18.42 & 28.48 & 33.38 & 23.55 & 20.55 & 82.99 & 14.82 & 17.15 & 17.63 & 22.44 & 20.48 & 26.45 & 52.56 \\
Ours & \textbf{87.26} & \textbf{22.75} & \textbf{33.10} & \textbf{29.95} & \textbf{31.67} & \textbf{38.91} & \textbf{29.75} &\textbf{ 26.91} & \textbf{83.74} & \textbf{19.93} & \textbf{27.56} & \textbf{29.72} & \textbf{34.87} & \textbf{30.29} & \textbf{36.94} & \textbf{62.06} \\
\midrule
\(\Delta_1\) & +0.95 & +9.59 & +3.57 & +10.37 & +4.97 & +4.68 & +8.00 & +7.18 & +0.57 & +12.89 & +21.87 & +22.39 & +19.70 & +13.70 & +21.53 & +10.31 \\
\(\Delta_2\) & +0.10 & +7.38 & +2.63 & +11.53 & +3.19 & +5.53 & +6.20 & +6.36 & +0.75 & +5.11 & +10.41 & +12.09 & +12.43 & +9.81 & +10.49 & +9.50 \\
\bottomrule
\end{tabular}
\raggedright
\footnotesize{\(\Delta_1\): The improvement compared with SegNext. \(\Delta_2\): The improvement compared with BEVFormer.}
\label{case2}
\end{table*}

\begin{table*}
\centering
\footnotesize
\caption{Fine-grained results of different models on the Brooklyn dataset. Except for BG in cross-view method, our method improves the mIoU by 2.27\% - 17.34\% and 3.92\% - 6.18\% for land use attribute, respectively, 0.64\% - 29.07\% and 0.22\% - 21.24\% for floor level attribute, respectively, compared with current state-of-the-art.}
\begin{tabular}{c@{\hspace{5pt}}c@{\hspace{5pt}}c@{\hspace{5pt}}c@{\hspace{5pt}}c@{\hspace{5pt}}c@{\hspace{5pt}}c@{\hspace{5pt}}c@{\hspace{5pt}}c@{\hspace{5pt}}c@{\hspace{5pt}}c@{\hspace{5pt}}c@{\hspace{5pt}}c@{\hspace{5pt}}c@{\hspace{5pt}}c@{\hspace{5pt}}c@{\hspace{5pt}}c}
\toprule
\multirow{3}{*}{Method} & \multicolumn{16}{c}{mIoU (\%) of each category} \\
\cmidrule(r){2-17}
 & \multicolumn{8}{c}{Land use} & \multicolumn{8}{c}{Floor} \\
\cmidrule(r){2-9} \cmidrule(r){10-17}
 & BG & C1 & C2 & C3 & C4 & C5 & C6 & C7 & BG & D1 & D2 & D3 & D4 & D5 & D6 & D7 \\
\midrule
SegNext \cite{guo2022segnext}& 83.62 & 35.32 & 35.22 & 35.17 & 30.41 & 22.09 & 39.96 & 12.97 & 84.73 & 31.92 & 34.92 & 34.51 & 29.47 & 5.29 & 31.25 & 27.28\\
BEVFormer \cite{li2022bevformer}& \textbf{86.28} & 48.74 & 42.76 & 39.78 & 31.68 & 25.39 & 48.68 & 20.37 & 84.67 & 38.93 & 46.73 & 45.00 & 44.23 & 21.48 & 40.59 & 35.11\\
Ours & 85.89 & \textbf{52.66} &\textbf{47.38} & \textbf{45.96} & \textbf{36.14} & \textbf{29.43} & \textbf{53.09} & \textbf{27.05}& \textbf{85.37} &\textbf{ 42.65} & \textbf{48.47} & \textbf{45.74} & \textbf{44.45} & \textbf{27.07} & \textbf{46.01} & \textbf{56.35} \\
\midrule
\(\Delta_1\) & +2.27 & +17.34 & +12.16 & +10.79 & +5.73 & +7.34 & +13.13 & +14.08 & +0.64 & +10.73 & +13.55 & +11.23 & +14.98 & +21.78 & +14.76 & +29.07 \\ 
\(\Delta_2\) & -0.39 & +3.92 & +4.62 & +6.18 & +4.46 & +4.04 & +4.41 & +6.68 & +0.70 & +3.72 & +1.74 & +0.74 & +0.22 & +5.59 & +5.42 & +21.24 \\
\bottomrule
\end{tabular}
\label{brooklyn}
\end{table*}

\begin{table*}[ht]
\centering
\footnotesize
\begin{minipage}[ht]{0.49\textwidth}
\centering
\caption{Fine-grained results of different models on the Boston dataset. Except for BG and E1 in cross-view method, our method improves the mIoU by 1.19\% - 13.59\% and 3.08\% - 4.24\% for land use attribute, respectively, compared with current state-of-the-art.}
\begin{tabular}{@{}c@{\hspace{9pt}}c@{\hspace{9pt}}c@{\hspace{9pt}}c@{\hspace{9pt}}c@{\hspace{9pt}}c@{\hspace{9pt}}c@{\hspace{9pt}}c@{}}
\toprule
\multirow{2}{*}{Method} & \multicolumn{7}{c}{mIoU (\%) of each category: Land use} \\
\cmidrule(r){2-8}
 & BG & E1 & E2 & E3 & E4 & E5 & E6 \\
\midrule
SegNext \cite{guo2022segnext}& 86.99 & 13.59 & 35.90 & 34.02 & 30.08 & 19.93 & 7.38 \\
BEVFormer \cite{li2022bevformer}& \textbf{88.24} & \textbf{15.97} & 39.85 & 37.05 & 37.75 & 29.28 & 11.96 \\
Ours & 88.18 & 15.90 & \textbf{44.08} & \textbf{40.40} &\textbf{ 40.83} & \textbf{33.52} & \textbf{15.16} \\
\midrule
\(\Delta_1\) & +1.19 & +2.31 & +8.18 & +6.38 & +10.75 & +13.59 & +7.78 \\
\(\Delta_2\) & -0.06 & -0.07 & +4.23 & +3.35 & +3.08 & +4.24 & +3.20 \\
\bottomrule
\end{tabular}
\label{Boston}
\end{minipage}
\hfill
\begin{minipage}[ht]{0.49\textwidth}
\centering
\caption{Fine-grained results of different methods on the Vigor dataset. Our method improves the mIoU by 1.30\% - 11.27\% and 0.24\% - 9.57\% for land use attribute, respectively, compared with current state-of-the-art.}
\begin{tabular}{@{}c@{\hspace{9pt}}c@{\hspace{9pt}}c@{\hspace{9pt}}c@{\hspace{9pt}}c@{\hspace{9pt}}c@{\hspace{9pt}}c@{}}
\toprule
\multirow{2}{*}{Method} & \multicolumn{6}{c}{mIoU (\%) of each category: Land use} \\
\cmidrule(r){2-7}
 & BG & F1 & F2 & F3 & F4 & F5 \\
\midrule
SegNext \cite{guo2022segnext}& 81.03 & 61.07 & 19.49 & 21.25 & 15.04 & 11.63 \\
BEVFormer \cite{li2022bevformer}& 82.09 & 63.51 & 25.74 & 23.86 & 16.55 & 12.25 \\
Ours & \textbf{82.33} & \textbf{67.65} & \textbf{30.76} & \textbf{28.76} & \textbf{26.12} & \textbf{14.54} \\
\midrule
\(\Delta_1\) & +1.30 & +6.58 & +11.27 & +7.51 & +11.08 & +2.91 \\
\(\Delta_2\) & +0.24 & +4.14 & +5.02 & +4.90 & +9.57 & +2.29 \\
\bottomrule
\end{tabular}
\label{Vigor}
\end{minipage}

\end{table*}


